\documentclass{article}

\usepackage[preprint]{corl_2026} 

\usepackage{wrapfig}
\usepackage{amsmath}
\usepackage{graphicx}
\usepackage{amssymb}
\usepackage{hyperref}
\usepackage{booktabs}
\usepackage{url}
\usepackage[table]{xcolor}
\usepackage{enumitem}
\usepackage{subcaption}
\usepackage{placeins}
\usepackage{tabularx}
\usepackage{array}
\usepackage{dsfont}
\usepackage{makecell}
\usepackage{float}

\newcolumntype{Y}{>{\raggedright\arraybackslash}X}

\title{TORL-VLA: Tactile Guided Online Reinforcement Learning for Contact-Rich Manipulation}

\newcommand{\BigName}[1]{{\bfseries #1}}

\author{
\BigName{Huaihang Zheng}\textsuperscript{1} \quad
\BigName{Yi Yang}\textsuperscript{1,4} \quad
\BigName{Kai Ma}\textsuperscript{1} \quad
\BigName{Shenglin Xu}\textsuperscript{1,2} \\
\BigName{Tian Xie}\textsuperscript{1} \quad
\BigName{Guozheng Li}\textsuperscript{1,5} \quad
\BigName{Xiangyu Wang}\textsuperscript{1,3} \quad
\BigName{Yiren Ma}\textsuperscript{1} \quad
\BigName{Si Liu}\textsuperscript{3} \\
\BigName{Yinian Mao}\textsuperscript{1} \quad
\BigName{Baoxu Liu}\textsuperscript{1} \\
\textsuperscript{1}Meituan \quad
\textsuperscript{2}Beijing Institute of Technology \quad
\textsuperscript{3}Beihang University \\
\textsuperscript{4}State Key Lab of Multimodal Artificial Intelligence Systems, Institute of Automation, CAS \\
\textsuperscript{5}China University of Mining and Technology (Beijing) \\
}

\begin{document}
\maketitle


\begin{abstract}
Vision-Language-Action (VLA) models have become a powerful framework for robotic manipulation, and recent studies have introduced tactile or force feedback into VLAs to address contact-rich tasks. However, these models are typically deployed as offline policies. When contact conditions shift from the training distribution, the policy cannot perform online adaptation, leading to problems such as inappropriate contact forces and inefficient retries. Therefore, we propose TORL-VLA, a tactile-guided online reinforcement learning framework that couples tactile feedback with policy refinement for contact-rich manipulation. Our method introduces a tactile-derived wrench-aware VLA to predict reference actions and future wrench sequences, while a lightweight online RL module is used to refine the reference actions. To stabilize learning from mixed exploratory policy-generated and human-intervention data, we introduce an intervention-censored critic that prevents post-intervention success from being wrongly credited to policy-generated actions preceding intervention. Real-robot experiments on long-horizon contact-rich tasks, including latch manipulation, coffee-cup placement, and egg handling, show that TORL-VLA improves success rates at both subtask and full-task levels, as well as time-bounded execution efficiency over strong baselines. Project page: \url{https://torl-vla.github.io/}.
\end{abstract}


\section{Introduction}

Vision-Language-Action (VLA) models \cite{zitkovich2023rt,o2024open,octo_2023,pmlr-v270-kim25c,black2024pi_0} have become an important paradigm for robotic manipulation by jointly modeling visual observations, language instructions, and large-scale robot trajectory data. However, many real-robot tasks also depend on fine-grained physical interaction states during contact. In tasks such as latch manipulation, cup placement, or fragile-object  handling, the robot must infer whether contact is sufficient, force is appropriate, or the object is slipping. These states are often weakly visible in RGB images, and the visual scene may appear close to success even when the underlying contact state has already deviated.

\begin{figure*}[t]
\centering
\includegraphics[
    width=\textwidth
]{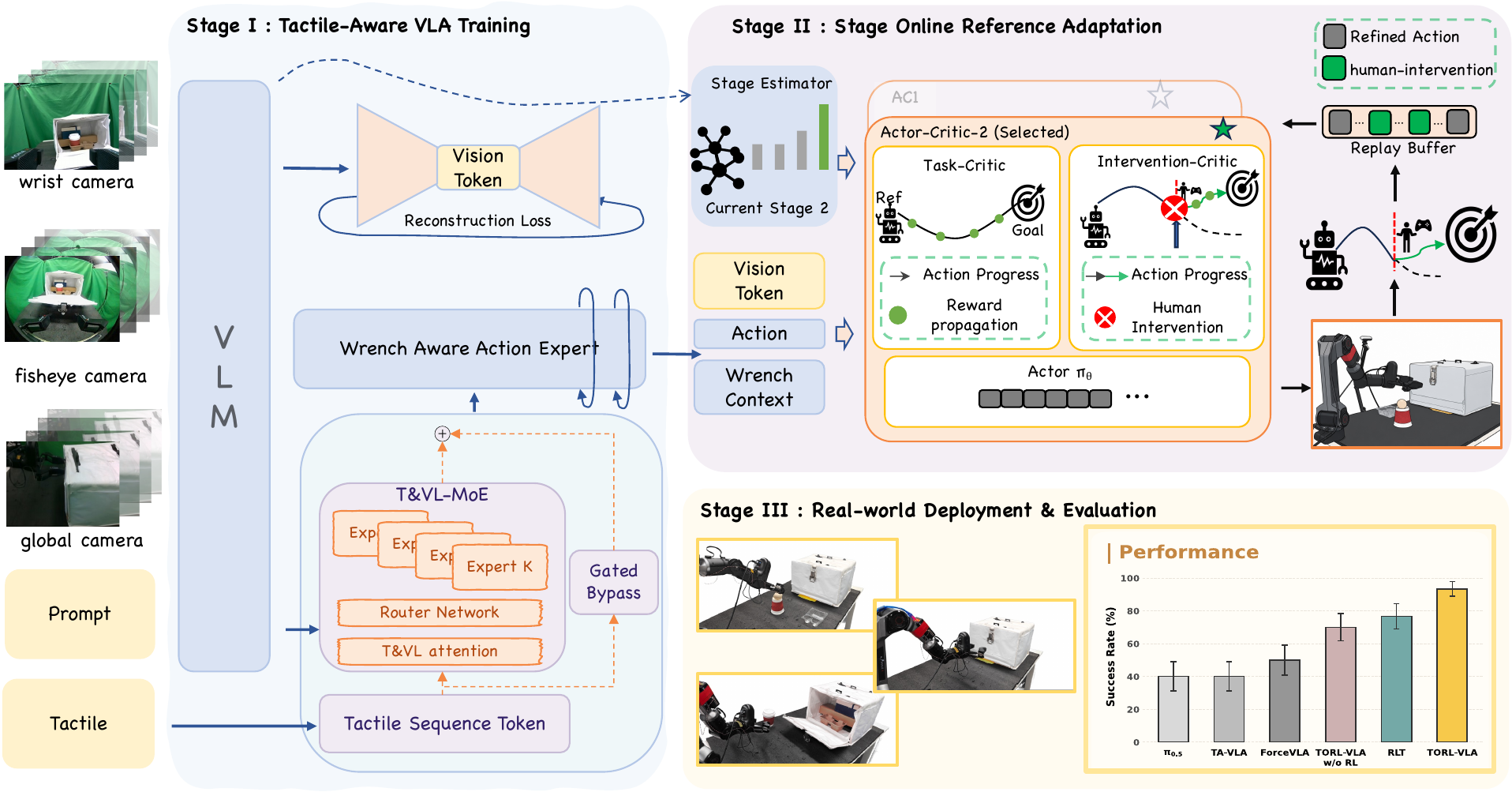}
\caption{\textbf{Overview of the TORL-VLA framework.} Stage \textbf{I}: Tactile-derived wrench sequences are fused after visual-language encoding through MoE routing to predict action references and future wrench sequences. Stage \textbf{II}: The frozen wrench-aware VLA provides vision tokens, reference actions, and predicted wrench; a lightweight actor-critic refines actions online with wrench conditioning and stage routing. Stage \textbf{III}: Evaluating performance on long-horizon contact-rich tasks.}
\label{fig:method_overview}
\vspace{-16pt}
\end{figure*}

Recent work has introduced tactile feedback into VLA models \cite{huang2025tactile,hao2026tla,zhang2025vtla,yu2026forcevla,li2026forcevla2,zhang2025ta,cheng2025omnivtla,bi2026vla,zhao2026fd,gubernatorov2026hapticvla}, allowing policies to use finer-grained physical interaction signals. These methods typically incorporate tactile signals as additional inputs, fusion features, or auxiliary prediction targets during supervised training, improving contact awareness. However, \textbf{contact awareness is not equivalent to contact adaptation}. Most tactile-aware VLAs remain static offline policies: once execution enters out-of-distribution contact states, they lack a mechanism to correct actions using current tactile feedback. Contact-rich manipulation therefore requires policies that can not only sense contact states, but also adapt offline action references online at key physical bottlenecks.

Based on this observation, we propose \textbf{TORL-VLA}, a \textbf{Tactile-Guided Online Reinforcement Learning} framework for contact-rich manipulation. TORL-VLA uses a tactile-derived wrench-aware VLA as an action-and-wrench reference model and refines its reference actions through a lightweight stage-specific online actor-critic module.
The actor generates executable action chunks conditioned on VLA reference actions, measured wrench feedback, predicted wrench cues, and compact visual-semantic tokens.
Through real-robot interaction, TORL-VLA learns contact-aware corrections around the VLA reference.

However, relying entirely on policy-generated exploration for online reinforcement learning~\cite{intelligence2025pi,chen2025conrft,yuanpolicy,xiao2026selfimproving,li2025gr,wagenmakersteering,xu2026rl} is costly and may introduce safety risks in real-world contact-rich tasks.
Human intervention~\cite{luo2024serl,lei2025rl,luo2025precise} is often needed when the robot is about to fail, becomes stuck, or produces unsafe contact.
This creates a value-learning bias: if a poor policy-generated action leads to human takeover and the rollout is later completed after intervention, a standard value function may wrongly credit that preceding policy action for the final success.

To address this issue, we introduce an \textbf{intervention-censored critic} in addition to the task critic.
It blocks post-intervention success from propagating back across human-correction boundaries, allowing TORL-VLA to use human corrections while avoiding over-crediting rescued outcomes to preceding policy-generated actions.

We validate TORL-VLA on real-robot long-horizon contact-rich manipulation tasks, including cup placement into a narrow holder, latch opening and closing, and fragile egg handling.
Compared with offline VLA execution, tactile/force-aware VLA baselines, and standard online refinement baselines, TORL-VLA improves success rates at both subtask and full-task levels, together with time-bounded execution efficiency.

In summary, our contributions are threefold:
\begin{enumerate}[leftmargin=2em]
    \item We propose \textbf{TORL-VLA}, a tactile-guided online reinforcement learning framework for contact-rich robotic manipulation. The framework uses a tactile-derived wrench-aware VLA as an action-and-wrench reference model and outputs refined action chunks through a lightweight online actor-critic module, turning a contact-aware VLA into a contact-adaptive policy.

    \item We introduce an \textbf{intervention-censored critic} for online RL. It addresses value-learning bias in human-intervened rollouts, enabling the policy to use human corrections, while avoiding over-crediting post-intervention success to preceding policy-generated actions.

    \item We validate TORL-VLA on real-robot long-horizon contact-rich manipulation tasks. Experiments show consistent improvements over strong baselines in success rates at both subtask and full-task levels, together with time-bounded execution efficiency.
\end{enumerate}

\section{Related Work}

\noindent\textbf{Tactile-Aware VLA.}
To improve perception and execution in contact-rich tasks, prior work has introduced tactile feedback into VLA policies. Tactile-VLA \cite{huang2025tactile}, TLA \cite{hao2026tla}, and VTLA \cite{zhang2025vtla} encode tactile images into tactile tokens and feed them together with visual-language features into the policy. ForceVLA \cite{yu2026forcevla}, ForceVLA2 \cite{li2026forcevla2}, and TA-VLA \cite{zhang2025ta} instead use lower-dimensional physical signals, injecting them into action generation so that the policy can exploit contact strength, force changes, and historical contact dynamics. OmniVTLA \cite{cheng2025omnivtla} and VLA-Touch \cite{bi2026vla} further study semantic alignment and dual-level tactile feedback for visual-language-action policies, while FD-VLA \cite{zhao2026fd} and HapticVLA \cite{gubernatorov2026hapticvla} use tactile or force distillation to exploit physical interaction signals during training and reduce sensor dependence at deployment. These methods show that tactile feedback can substantially improve the contact awareness of VLA policies. However, most of them are still trained under supervised learning or offline imitation learning and deployed as fixed policies. In contrast, we focus on using tactile feedback to adaptively correct VLA actions online, rather than only enhancing contact awareness during offline training.

\noindent\textbf{Online RL for VLA.}
Applying RL to VLA policies mainly aims to go beyond the performance ceiling of offline demonstrations. RECAP \cite{intelligence2025pi} learns from demonstrations, on-policy rollouts, and human intervention data with an advantage-conditioned policy; ConRFT \cite{chen2025conrft} uses a consistency policy for reinforced fine-tuning; Policy Decorator \cite{yuanpolicy} and PLD \cite{xiao2026selfimproving} learn residual corrections on top of existing policies; GR-RL \cite{li2025gr} and DSRL \cite{wagenmakersteering} apply RL in the latent or noise space of generative action models to steer the action-generation process; and RLT \cite{xu2026rl} connects a VLA to an online actor-critic through an RL token, enabling local policy improvement around VLA reference actions. SERL \cite{luo2024serl} and RL100 \cite{lei2025rl} also show that off-policy actor-critic methods combined with demonstrations and human corrections can improve contact manipulation policies with limited real-robot data. These works show that RL can improve VLA or robot policy performance, but they rarely address how human intervention can bias value learning when post-intervention success is attributed to preceding policy-generated actions. In contrast, TORL-VLA introduces intervention-censored value learning to block such misleading credit propagation across human-correction boundaries.

\section{Method}
\label{sec:method}

TORL-VLA aims to turn an offline VLA with tactile-derived wrench feedback into a contact-adaptive policy for real-robot manipulation, as illustrated in Fig.~\ref{fig:method_overview}.
The wrench-aware reference model predicts a reference action chunk together with a future wrench sequence, providing both a motion prior and an expected contact reference. A lightweight online RL refiner then conditions on these references
to generate corrected executable actions.


\subsection{Tactile-Aware VLA as an Action-and-Wrench Reference Model}

\begin{wrapfigure}{r}{0.52\textwidth}
\vspace{-10pt}
\centering
\includegraphics[width=0.50\textwidth]{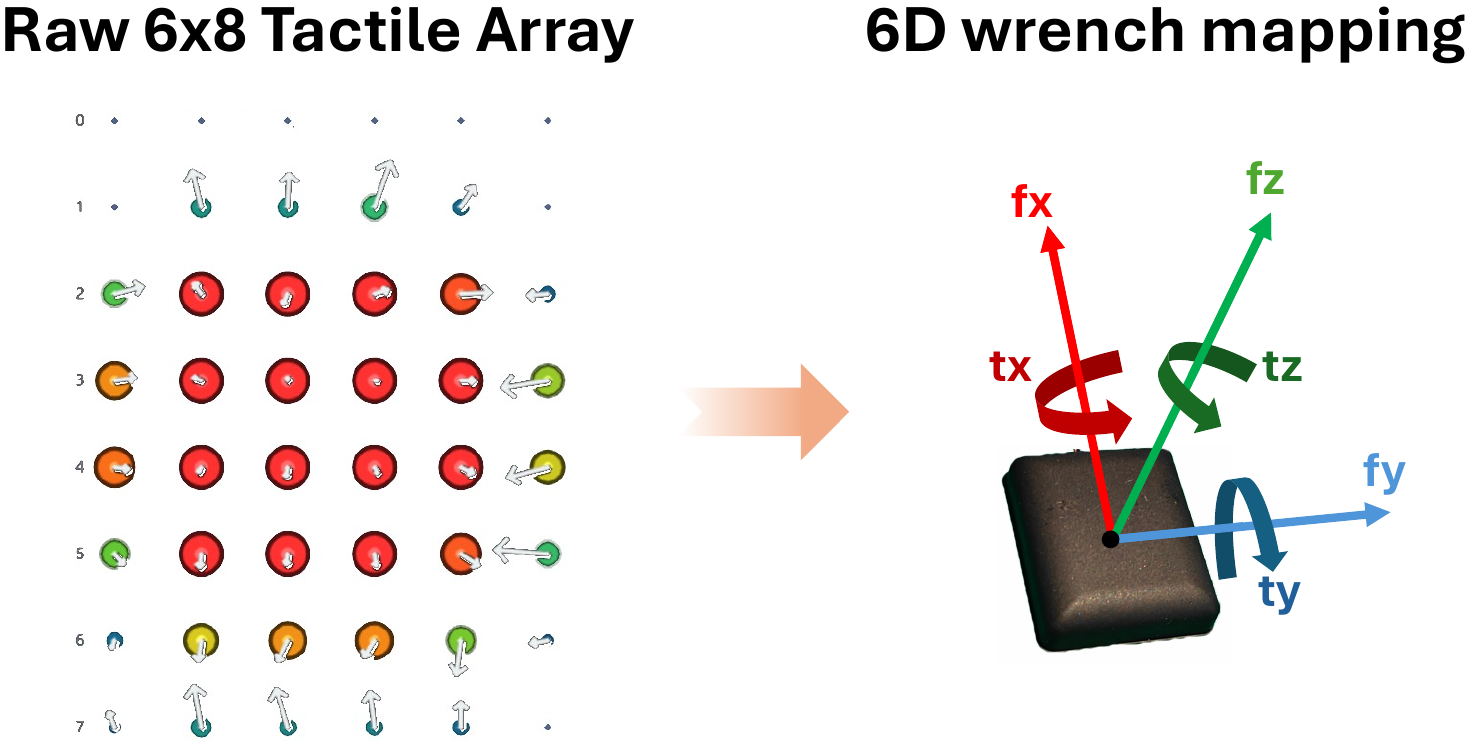}
\caption{\textbf{Tactile-to-wrench mapping.}
Tactile readings are mapped to a 6D wrench representation for contact-aware manipulation.}
\label{fig:tactile_wrench_mapping}
\vspace{-8pt}
\end{wrapfigure}

\paragraph{Tactile-derived wrench representation.}
Our gripper has two piezoelectric tactile pads mounted on the inner finger surfaces, as illustrated in Fig.~\ref{fig:tactile_wrench_mapping}. Each raw \(6 \times 8\) tactile array is mapped to a compact 6D wrench through the calibrated sensor interface. This wrench representation provides a compact physical interface and facilitates reuse with tactile sensors that can be mapped to a common force-torque space. For fingertip \(i \in \{L, R\}\), the tactile-derived wrench at timestep \(t\) is \(\mathbf{w}_t^i =
    [f_x^i, f_y^i, f_z^i, \tau_x^i, \tau_y^i, \tau_z^i]
    \in \mathbb{R}^{6}\), 
where \(f^i\) and \(\tau^i\) denote the three-axis force and torque components, respectively. The full tactile observation is \(\mathbf{w}_t=[\mathbf{w}_t^L,\mathbf{w}_t^R]\in\mathbb{R}^{12}\).

\paragraph{Temporal wrench encoding.}
To capture short-term contact dynamics, we encode a recent wrench sequence \(\mathbf{W}_t=\{\mathbf{w}_{t-J+1},\ldots,\mathbf{w}_t\}\), where \(J\) is the temporal window length, into a single wrench token \(\mathbf{z}_t^{\mathrm{tac}}\in\mathbb{R}^{1\times d_{\mathrm{vl}}}\) using a lightweight MLP encoder. This fixed-size token allows different wrench sequence lengths to be fused without changing the subsequent architecture.

\paragraph{Wrench-conditioned reference generation.}
The VLM encodes the image, language, and robot-state prefix tokens into hidden states \(\mathbf{P}_t \in \mathbb{R}^{N_p \times d_{\mathrm{vl}}}\), where \(N_p\) is the number of prefix tokens and \(d_{\mathrm{vl}}\) is the hidden dimension of the VLM stream \cite{intelligence2025pi05}. At flow-matching interpolation time \(s\in[0,1]\), the action expert processes the noisy generation variable \(\mathbf{x}_{t,s}^{\mathrm{fm}}\) and produces action-side hidden states \(\mathbf{S}_{t,s} \in \mathbb{R}^{H \times d_{\mathrm{ae}}}\), where \(H\) is the prediction horizon and \(d_{\mathrm{ae}}\) is the action-expert hidden dimension.

Inspired by prior multimodal fusion designs \cite{alayrac2022flamingo,li2023blip,driess2026knowledge,hancock2025actions}, rather than feeding the tactile-derived wrench token into the VLM with image and language tokens, we fuse it after visual-language encoding by applying self-attention to \([\mathbf{P}_t;\mathbf{z}_t^{\mathrm{tac}}]\), yielding \(\mathbf{H}_t \in \mathbb{R}^{(N_p+1)\times d_{\mathrm{vl}}}\). 

This preserves pretrained visual-language semantics while enabling interaction with tactile-derived wrench feedback. Following recent MoE-based fusion and adaptation designs \cite{yu2026forcevla,miao2025fedvla,shen2025expertise,du2025himoe}, a soft router computes \(\boldsymbol{\rho}_t=\mathrm{Softmax}(R_{\mathrm{tac}}(\mathbf{H}_t))\) and produces the fused feature \(\mathbf{F}_t=\sum_{m=1}^{M}\boldsymbol{\rho}_{t,m}\odot E_m(\mathbf{H}_t)\), where \(E_m\) is the \(m\)-th expert and \(\odot\) denotes broadcasted element-wise weighting.

To modulate action generation, the routed feature is projected to the action horizon and combined with a zero-initialized wrench-token bypass:
\begin{equation}
    \mathbf{G}_t =
    \Pi_{\mathrm{tac}}(\mathbf{F}_t)
    +
    \alpha P_{\mathrm{tac}}^{H}(\mathbf{z}_t^{\mathrm{tac}})
    \in \mathbb{R}^{H \times d_{\mathrm{ae}}},
\end{equation}
where \(\Pi_{\mathrm{tac}}\) maps the routed fusion feature to the action horizon, \(P_{\mathrm{tac}}^{H}\) projects and expands the wrench token to \(H\) horizon-aligned guidance tokens, and \(\alpha\) is a learnable scalar initialized to zero. 
The zero-initialized gate adaptively controls the bypass contribution, yielding the wrench-conditioned action-side representation \(\tilde{\mathbf{S}}_{t,s}=\mathbf{S}_{t,s}+\mathbf{G}_t\).

\paragraph{Joint action-and-wrench prediction.}
The output head jointly predicts \(\hat{\mathbf{y}}_t=[\hat{\mathbf{A}}_{t:t+H-1}, \hat{\mathbf{W}}_{t:t+H-1}]\), where \(\hat{\mathbf{A}}_{t:t+H-1}\) is the predicted action chunk and \(\hat{\mathbf{W}}_{t:t+H-1}\) is the co-predicted future wrench sequence \cite{li2026forcevla2,zhang2025ta,li2026favla,ye2025learning}. The action chunk provides the motion reference, while the wrench sequence provides expected contact guidance for learning action--contact interaction dynamics.

We train the reference model with a joint action-wrench flow matching objective over the ground-truth target \(\mathbf{y}_t=[\mathbf{A}_{t:t+H-1}, \mathbf{W}_{t:t+H-1}]\). Following the standard flow-matching formulation, the model predicts the velocity field \(\hat{\mathbf{u}}_{t,s}\), and the loss is separated into action and wrench dimensions:
\begin{equation}
    \mathcal{L}_{\mathrm{ref}}
    =
    \left\|
    \hat{\mathbf{u}}^{a}_{t,s} - \mathbf{u}^{a}_{t,s}
    \right\|_2^2
    +
    \lambda_{w}
    \left\|
    \hat{\mathbf{u}}^{w}_{t,s} - \mathbf{u}^{w}_{t,s}
    \right\|_2^2,
\end{equation}
where the superscripts \(a\) and \(w\) denote the action and wrench dimensions, and \(\lambda_w\) balances action and wrench prediction.

\subsection{Stage-Specific Online Reference Adaptation}
\label{sec:online_contact_adaptation}

The wrench-aware reference model described above jointly predicts action references and future wrench sequences, providing both semantic and physical priors for deployment.
However, offline VLA policies can still fail at contact-rich bottlenecks due to local pose errors, friction variation, object compliance, or accumulated execution error, motivating online reinforcement learning for local refinement~\cite{black2024pi_0,intelligence2025pi,yuanpolicy,xiao2026selfimproving,xu2026rl,intelligence2025pi05}.
We therefore design a stage-specific online actor-critic framework that refines VLA action references using measured wrench feedback and predicted future wrench cues.
A stage estimator routes each contact-rich phase to a corresponding actor-critic head, while an intervention-censored value objective improves credit assignment under human-intervened rollouts.

\paragraph{Stage-specific wrench-conditioned action refinement.}
At each policy query, the frozen wrench-aware VLA predicts an \(H\)-step action sequence \(\hat{\mathbf{A}}_{t:t+H-1}\) and an \(H\)-step future wrench sequence \(\hat{\mathbf{W}}_{t:t+H-1}\).
The robot executes only the first \(K\le H\) steps before replanning.
We denote the executable VLA action reference and the aligned predicted wrench cue as \(\mathbf{A}_t^{\mathrm{ref}}=\hat{\mathbf{A}}_{t:t+K-1}\) and \(\hat{\mathbf{W}}_t^{\mathrm{ref}}=\hat{\mathbf{W}}_{t:t+K-1}\), respectively.
The online actor uses the refinement context \(\mathbf{c}_t=[\mathbf{z}_t^{\mathrm{vla}};\mathbf{q}_t;\mathbf{w}_t;\mathbf{W}_{t-L:t-1};\hat{\mathbf{W}}_t^{\mathrm{ref}};\mathbf{A}_t^{\mathrm{ref}}]\), where \(\mathbf{z}_t^{\mathrm{vla}}\) denotes compact visual-semantic tokens extracted from the frozen VLA, \(\mathbf{q}_t\) is robot proprioception, \(\mathbf{w}_t\) is the current measured wrench, and \(\mathbf{W}_{t-L:t-1}\) is the recent measured wrench history.
A stage estimator predicts the current contact stage \(g_t\in\{1,\ldots,G\}\) and selects the corresponding stage-specific actor head \(\pi_{\phi_{g_t}}\).
The selected actor head outputs the refined executable action chunk \(\mathbf{A}_t^{\phi}=\pi_{\phi_{g_t}}(\mathbf{c}_t)\).

\paragraph{Intervention-censored actor-critic learning.}
The stage-specific actor-critic heads are trained from rollouts containing both policy-generated actions and human-intervention actions.
Under sparse task rewards, a policy-generated action chunk may lead to an abnormal contact state that requires human correction; if the rollout is later completed after intervention, a standard task critic may over-credit the preceding policy-generated chunk~\cite{luo2025precise,ross2011reduction}.
To address this credit-assignment issue, we use a task critic \(Q_{\mathrm{task}}(\mathbf{c}_t,\mathbf{A}_t)\) to estimate the expected task return of an evaluated action chunk \(\mathbf{A}_t\), and introduce an additional intervention-censored critic \(Q_{\mathrm{ic}}(\mathbf{c}_t,\mathbf{A}_t)\) to prevent post-intervention success from being propagated back across human-correction boundaries.

We identify such boundaries using the control source stored in replay.
For each executed \(K\)-step chunk \(\mathbf{A}_t^{\mathrm{data}}=(\mathbf{a}_{t,0}^{\mathrm{data}},\ldots,\mathbf{a}_{t,K-1}^{\mathrm{data}})\), replay stores a label \(\sigma_{t,j}\in\{\mathrm{policy},\mathrm{human}\}\) for each step.
Let \(h_t\) denote the fraction of human-intervention steps in chunk \(t\).
Given a threshold \(\rho_{\mathrm{int}}\), the intervention-boundary mask \(m_t^{\mathrm{int}}\) is set to one when a mostly policy-generated chunk is followed by an intervention-dominated chunk, and zero otherwise.
This mask is used only to censor value bootstrapping; intervention actions remain in replay for behavior regularization and task-return learning.

Let \(R_t^K=\sum_{j=0}^{K-1}\gamma^j r_{t,j}\) be the discounted chunk reward, \(d_t\in\{0,1\}\) be the terminal flag, and \(\bar{\mathbf{A}}_{t+K}\) be the target action chunk for bootstrapping at the next context \(\mathbf{c}_{t+K}\).
The task critic uses the standard target \(y_{\mathrm{task}}=R_t^K+(1-d_t)\gamma^K\min_{i=1,2}Q_{\mathrm{task}}^{(i)}(\mathbf{c}_{t+K},\bar{\mathbf{A}}_{t+K})\)~\cite{fujimoto2018addressing,haarnoja2018soft}, while the intervention-censored critic uses
\begin{equation}
y_{\mathrm{ic}}
=
R_t^K
-
c_{\mathrm{int}}m_t^{\mathrm{int}}
+
(1-d_t)(1-m_t^{\mathrm{int}})\gamma^K
\min_{i=1,2}
Q_{\mathrm{ic}}^{(i)}
\left(
\mathbf{c}_{t+K},
\bar{\mathbf{A}}_{t+K}
\right),
\label{eq:ic_target}
\end{equation}
where \(c_{\mathrm{int}}\) is the intervention cost.
When \(m_t^{\mathrm{int}}=1\), the bootstrap term is removed and the intervention cost is applied, so success achieved after human correction is not assigned to the preceding policy-generated chunk.

The actor is optimized with behavior regularization and value-based improvement.
For behavior regularization, policy-generated steps are anchored to the VLA reference, while human-intervention steps imitate the corrected executed action.
This gives \(\mathcal{L}_{\mathrm{bc}}=\sum_{j=0}^{K-1}\|\mathbf{a}_{t,j}^{\phi}-\mathbf{a}_{t,j}^{\mathrm{bc}}\|_2^2\), where \(\mathbf{a}_{t,j}^{\mathrm{bc}}\) is set according to the recorded control source.
For value-based improvement, we compare the actor output with the VLA reference using \(\Delta Q_{\mathrm{ic}}=[\min_i Q_{\mathrm{ic}}^{(i)}(\mathbf{c}_t,\mathbf{A}_t^{\mathrm{ref}})-\min_i Q_{\mathrm{ic}}^{(i)}(\mathbf{c}_t,\mathbf{A}_t^\phi)]_+\), where \([x]_+=\max(x,0)\).
The actor objective is
\begin{equation}
\mathcal{L}_{\pi}
=
\lambda_{\mathrm{bc}}\mathcal{L}_{\mathrm{bc}}
-
\lambda_{\mathrm{task}}
\mathbb{E}
\left[
\min_{i=1,2}
Q_{\mathrm{task}}^{(i)}
\left(
\mathbf{c}_t,
\mathbf{A}_t^{\phi}
\right)
\right]
+
\lambda_{\mathrm{ic}}
\mathbb{E}
\left[
\Delta Q_{\mathrm{ic}}
\right].
\label{eq:actor_objective}
\end{equation}
The task critic encourages task-return improvement, the behavior term keeps the actor close to the VLA reference or human correction according to the control source, and the IC-value penalty discourages refinements with lower intervention-censored value than the VLA reference.

\begin{wrapfigure}{R}{0.49\columnwidth}
\vspace{-8pt}
\centering
\includegraphics[width=\linewidth]{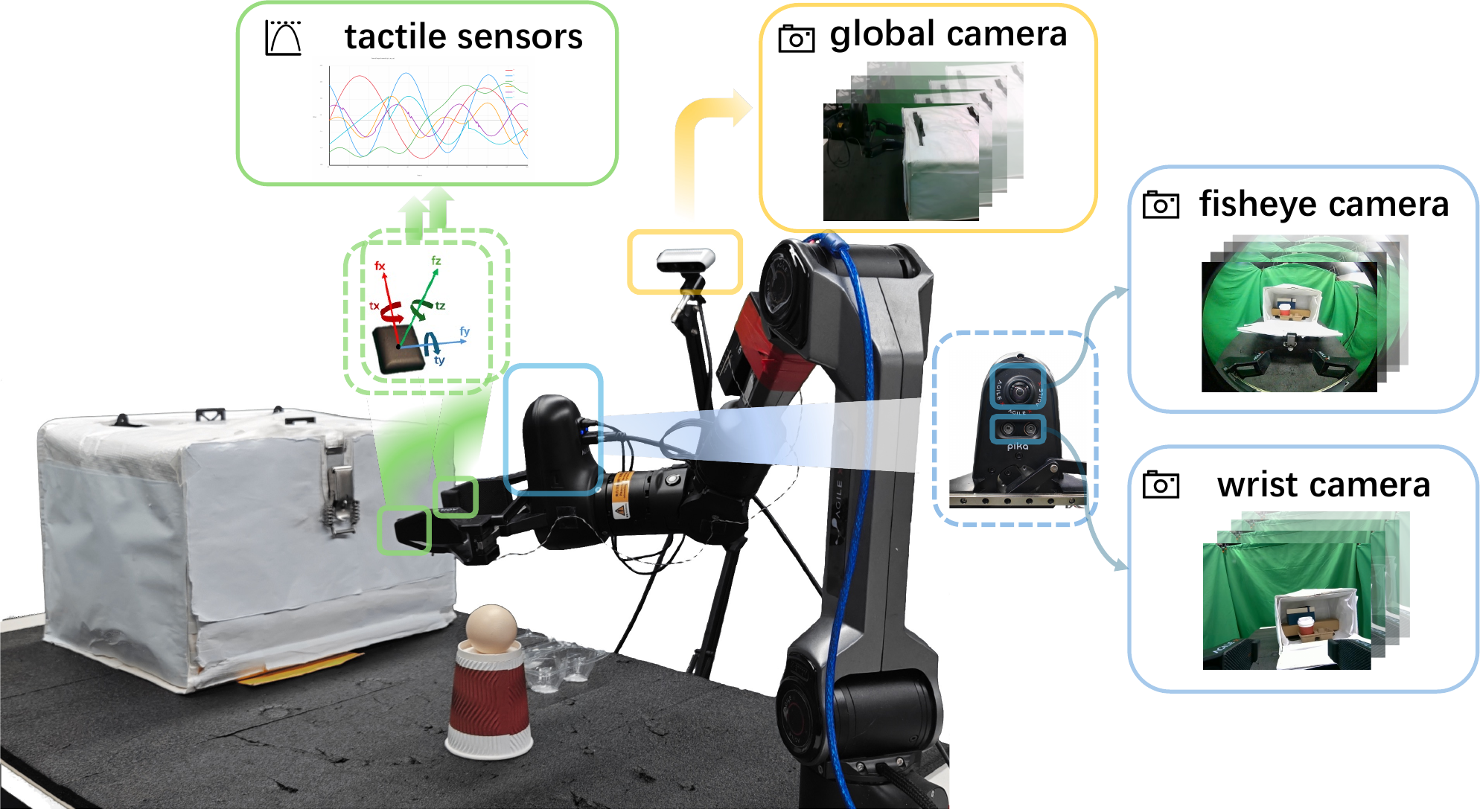}
\caption{\textbf{Experimental platform.}
Real-robot latch-box setup with multi-view vision and dual-fingertip tactile sensing.}
\label{fig:experimental_framework}
\vspace{-8pt}
\end{wrapfigure}

\section{Experiments}
\label{sec:experiments}

We evaluate TORL-VLA on contact-rich manipulation tasks to answer three questions:
(i) whether wrench-guided online refinement improves subtask and full-task success;
(ii) whether the wrench-aware reference model provides stronger action-and-wrench references than adapted physical-feedback VLA baselines; and
(iii) whether wrench-context conditioning and intervention-censored value learning are necessary for online adaptation.
We report subtask success, full-task success, 60-min full-task throughput, and ablations of the reference model and online RL components. Additional experimental details are provided in the Appendix, and the code will be released for reference.

\subsection{Experimental Setup}

\begin{figure*}[!t]
\vspace{-4pt}
\centering
\includegraphics[width=\textwidth]{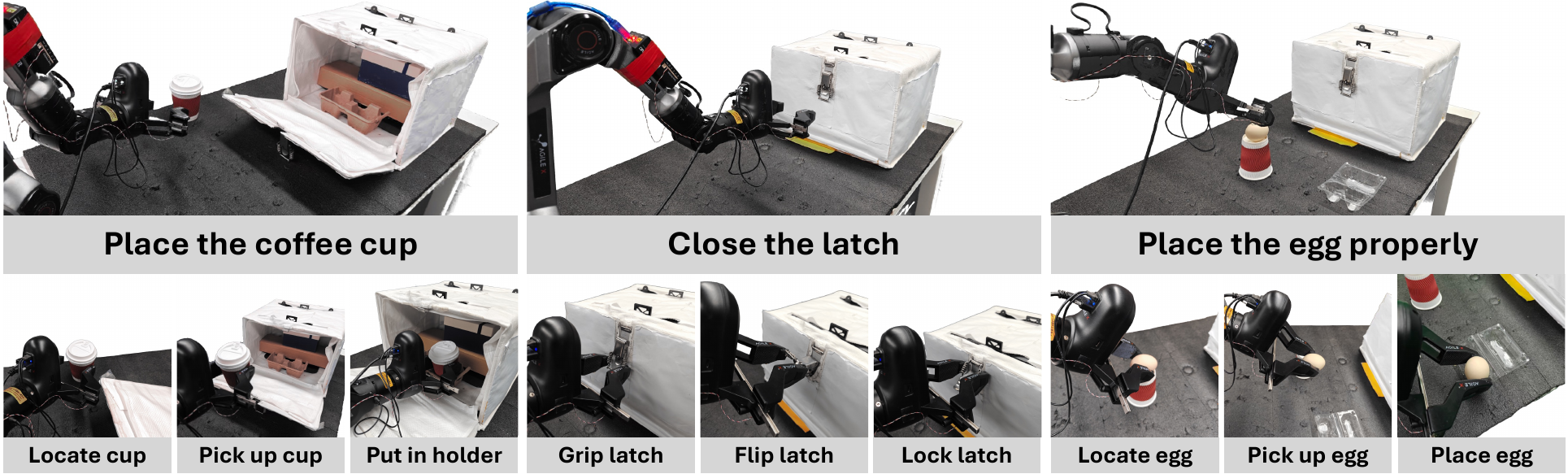}
\caption{\textbf{Task visualization.} Representative contact-rich subtasks in latch-box manipulation, including coffee-cup placement, latch locking, and egg placement. } \label{fig:experiment_visualization}
\vspace{-8pt}
\end{figure*}

\paragraph{Tasks and Platform.}
We evaluate TORL-VLA on a real-robot latch-box setup with three contact-centric subtasks: Coffee Cup, Latch, and Egg.
Coffee Cup requires placing a cup into a tight holder inside the box with limited visibility. Latch requires gripping, flipping, and locking a mechanical latch, and Egg requires fragile-object grasping and placement.
These tasks cover tight insertion, mechanical engagement, and fragile-object handling, where vision alone is often insufficient to infer the contact state.
The platform and task stages are shown in Fig.~\ref{fig:experimental_framework} and Fig.~\ref{fig:experiment_visualization}.

\paragraph{Evaluation Protocol.}
We report 
subtask success, full-task success, and full-task throughput.
Subtask and full-task success rates are evaluated over 30 autonomous trials.
Full-task evaluation starts from the same initial latch-box state and requires completing all stages in one autonomous rollout without intervention, reset, or safety stop. To assess time-bounded execution efficiency, we also report 60-min throughput, defined as the number of successful full-task completions within a fixed 60-minute evaluation window. For ablations, reference-model variants are evaluated on the three contact-centric subtasks, while online-adaptation variants are compared by final subtask success and adaptation curves under the 
same accumulated online robot data.

\begin{wrapfigure}{r}{0.47\columnwidth}
\vspace{-8pt}
\centering
\includegraphics[width=\linewidth]{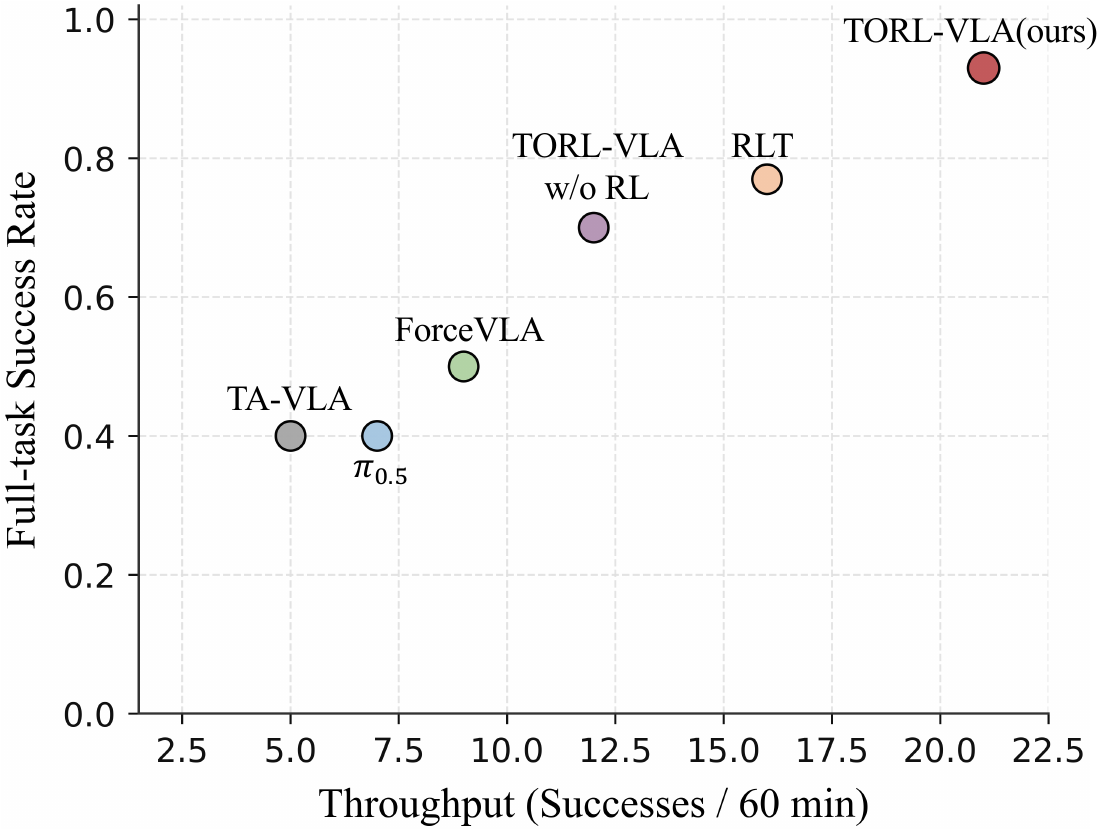}
\caption{\textbf{Full-task reliability and throughput.}
Success rate and 60-min throughput are compared across methods.}
\label{fig:fulltask_success_vs_throughput}
\vspace{-10pt}
\end{wrapfigure}

\paragraph{Baselines and Fairness.}
We compare TORL-VLA with $\pi_{0.5}$, TA-VLA, ForceVLA, TORL-VLA w/o RL, and RLT under the same robot platform, action representation, demonstrations, execution protocol, and evaluation trials.
TA-VLA and ForceVLA are reimplemented on the same $\pi_{0.5}$ backbone and adapted to the same dual-fingertip wrench interface.
TORL-VLA w/o RL directly executes the learned wrench-aware reference model, while RLT uses standard reference-guided online refinement without the intervention-censored critic. All methods are evaluated using the same human-judged binary success signal for each subtask, which also serves as the sparse reward when training the online RL module.

\subsection{Performance on Contact-Rich Tasks}

Table~\ref{tab:real_robot_results} compares TORL-VLA with offline VLA baselines, tactile-aware VLA variants, and online refinement baselines.

\begin{table}[!t]
\centering
\caption{\textbf{Real-robot performance.}
Success rates are reported for three contact-centric subtasks and the full task; FT Avg. Time is computed over successful full-task trials.}
\label{tab:real_robot_results}
\footnotesize
\renewcommand{\arraystretch}{1.05}
\setlength{\tabcolsep}{2.6pt}

\newcolumntype{L}[1]{>{\raggedright\arraybackslash}p{#1}}
\newcolumntype{C}[1]{>{\centering\arraybackslash}p{#1}}

\resizebox{\columnwidth}{!}{%
\begin{tabular}{L{0.22\columnwidth}
                C{0.100\columnwidth}
                C{0.100\columnwidth}
                C{0.100\columnwidth}
                !{\vrule width 0.3pt}
                C{0.170\columnwidth}
                C{0.210\columnwidth}}
\toprule
Method
& Cup $\uparrow$
& Latch $\uparrow$
& Egg $\uparrow$
& FT Success $\uparrow$
& FT Avg. Time (s) $\downarrow$ \tabularnewline
\midrule
$\pi_{0.5}$
& 18/30 & 15/30 & 20/30 & 12/30 & 199.65 \tabularnewline
TA-VLA
& 19/30 & 17/30 & 20/30 & 12/30 & 204.45 \tabularnewline
ForceVLA
& 21/30 & 20/30 & 22/30 & 15/30 & 195.34 \tabularnewline
TORL-VLA w/o RL
& 25/30 & 23/30 & 25/30 & 21/30 & 191.91 \tabularnewline
RLT
& 26/30 & 25/30 & 25/30 & 23/30 & 175.23 \tabularnewline
\midrule
\rowcolor{gray!15}
\textbf{TORL-VLA}
& \textbf{30/30}
& \textbf{29/30}
& \textbf{30/30}
& \textbf{28/30}
& \textbf{165.45} \tabularnewline
\bottomrule
\end{tabular}%
}
\vspace{-6pt}
\end{table}

The base $\pi_{0.5}$ policy struggles when subtle contact states, such as partial insertion, latch engagement, or unstable grasping, cannot be reliably inferred from vision alone.
TA-VLA yields marginal local gains, and ForceVLA improves several bottlenecks more clearly; however, both remain offline policies and cannot actively correct contact deviations during execution. 
TORL-VLA w/o RL improves over these offline baselines, indicating that the wrench-aware reference model provides stronger action-and-wrench references.
However, directly executing the reference remains insufficient for handling real contact shifts during long-horizon execution.
RLT further improves local contact handling, but its full-task success remains below TORL-VLA.
In the egg task, 3 of the 5 RLT failures are caused by overly aggressive grasps that break the egg, whereas TORL-VLA w/o RL mainly fails due to imprecise grasping positions.
TORL-VLA achieves near-perfect subtask success and improves full-task completion from 12/30 with $\pi_{0.5}$ to 28/30.
Fig.~\ref{fig:fulltask_success_vs_throughput} compares full-task reliability and time-bounded productivity.
TORL-VLA achieves the best result on both metrics, suggesting more reliable and efficient long-horizon execution under contact deviations.

\subsection{Ablation Studies}


\begin{table*}[!t]
\centering
\caption{\textbf{Ablation studies.}
Left: wrench-aware reference model ablation. Right: online adaptation ablation. IC Critic denotes the intervention-censored critic.}
\label{tab:ablation_studies}
\footnotesize
\renewcommand{\arraystretch}{0.98}
\setlength{\tabcolsep}{3.0pt}

\newcolumntype{A}{>{\raggedright\arraybackslash}X}
\newcolumntype{B}{>{\centering\arraybackslash}p{0.16\linewidth}}

\begin{minipage}[t]{0.48\textwidth}
\centering
\textbf{Wrench-aware reference model ablation}
\vspace{1pt}

\begin{tabularx}{\linewidth}{A B B B}
\toprule
Configuration
& Cup $\uparrow$
& Latch $\uparrow$
& Egg $\uparrow$ \tabularnewline
\midrule
w/o WHT
& 24/30 & 22/30 & 22/30 \tabularnewline
w/o FWP
& \textbf{25/30} & 21/30 & 24/30 \tabularnewline
w/o MoE
& 18/30 & 17/30 & 19/30 \tabularnewline
w/o Bypass
& 23/30 & 20/30 & 21/30 \tabularnewline
\midrule
\rowcolor{gray!15}
\textbf{Full w/o RL}
& \textbf{25/30} & \textbf{23/30} & \textbf{25/30} \tabularnewline
\bottomrule
\end{tabularx}
\end{minipage}
\hfill
\begin{minipage}[t]{0.48\textwidth}
\centering
\textbf{Online adaptation ablation}
\vspace{1pt}

\begin{tabularx}{\linewidth}{A B B B}
\toprule
Configuration
& Cup $\uparrow$
& Latch $\uparrow$
& Egg $\uparrow$ \tabularnewline
\midrule
w/o Wrench Context
& 27/30 & 27/30 & 26/30 \tabularnewline
w/o IC Critic
& 27/30 & 26/30 & 28/30 \tabularnewline
\midrule
\rowcolor{gray!15}
\textbf{Full}
& \textbf{30/30} & \textbf{29/30} & \textbf{30/30} \tabularnewline
\bottomrule
\end{tabularx}
\end{minipage}

\vspace{-8pt}
\end{table*}

\paragraph{Wrench-Aware Reference Model.}
We ablate the wrench history token (WHT), future wrench prediction objective (FWP), mixture-of-experts (MoE) fusion module, and zero-initialized physical bypass (Bypass). The results are summarized in Table~\ref{tab:ablation_studies}. 
Removing MoE causes the largest degradation, highlighting the importance of routing wrench features with visual-language representations.
Removing Bypass also reduces success, especially for Latch and Egg, suggesting that the direct wrench-token bypass preserves fine-grained contact information.
In contrast, removing WHT or FWP has smaller effects, indicating that temporal wrench encoding and future wrench prediction mainly complement the fusion pathway.
The Full w/o RL variant achieves the best overall reference-model performance.

\begin{figure*}[!t]
\centering
\begin{subfigure}[t]{0.48\textwidth}
\centering
\includegraphics[width=\linewidth]{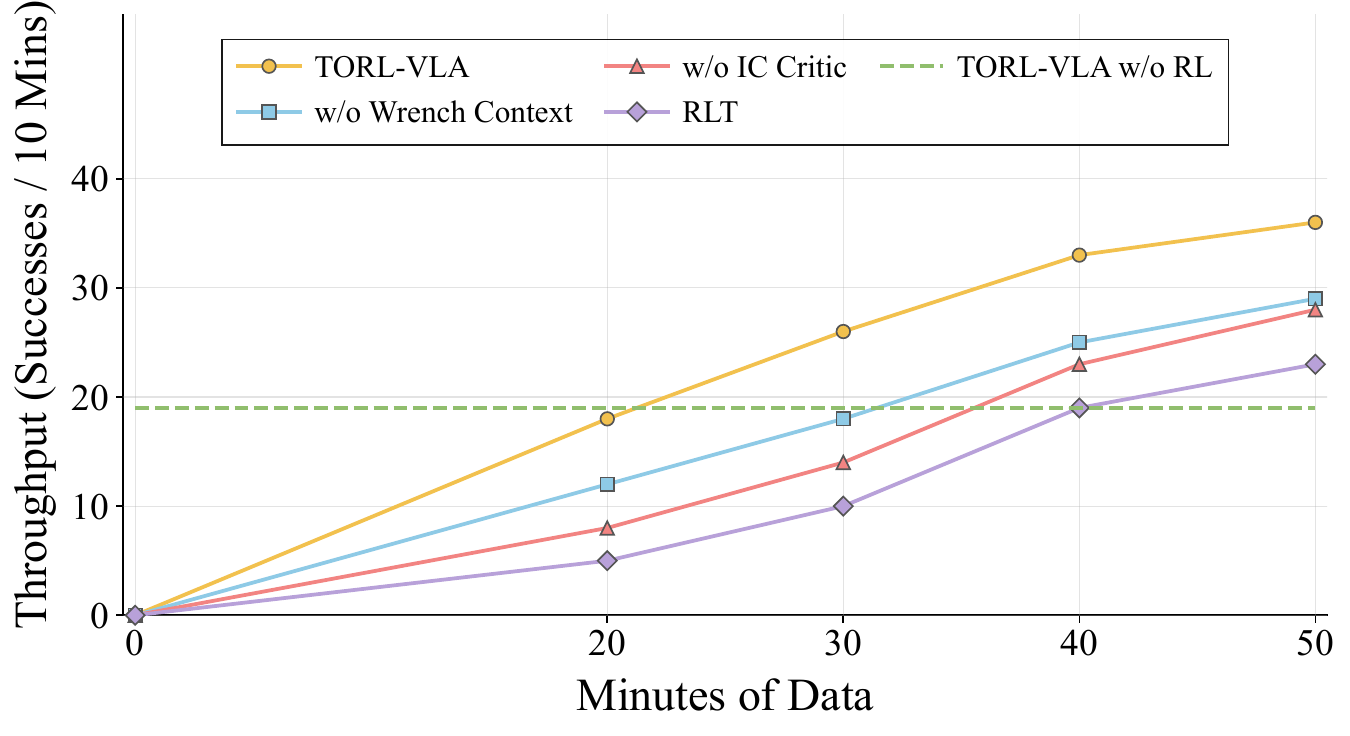}
\caption{Throughput.}
\label{fig:throughput_curve}
\end{subfigure}
\hfill
\begin{subfigure}[t]{0.48\textwidth}
\centering
\includegraphics[width=\linewidth]{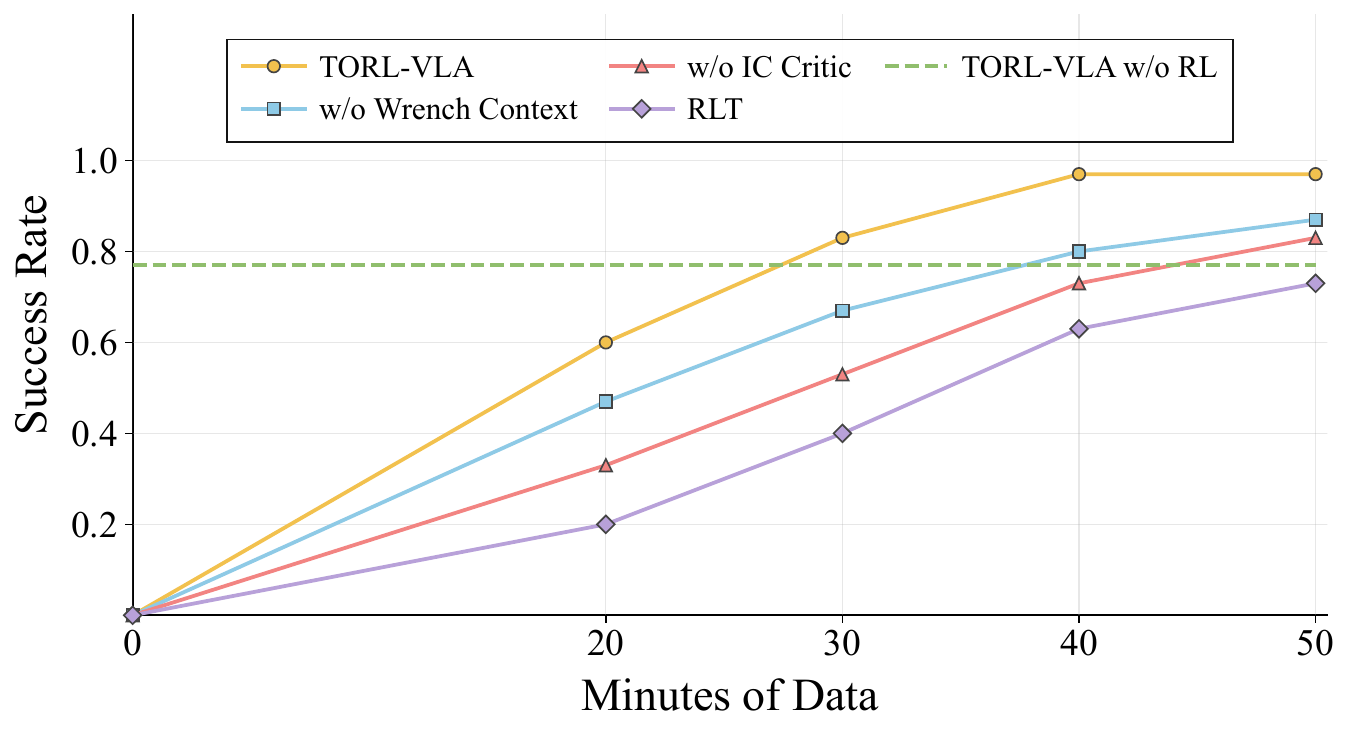}
\caption{Success rate.}
\label{fig:success_rate_curve}
\end{subfigure}
\caption{\textbf{Online adaptation on Latch.}
Left: throughput measured per 10 minutes as online data increases.
Right: success rate at each data checkpoint.}
\label{fig:online_adaptation_curves}
\vspace{-6pt}
\end{figure*}

\paragraph{Online Adaptation Ablation.}
Table~\ref{tab:ablation_studies} and Fig.~\ref{fig:online_adaptation_curves} analyze the online refinement module from final performance and adaptation speed, respectively.
All variants use the same frozen wrench-aware reference model and stage-specific actor-critic structure, but differ in the refinement context or value-learning objective.
Removing wrench context reduces success across all subtasks, indicating that measured wrench feedback and predicted wrench cues provide useful state information for contact-aware action refinement.
Removing the intervention-censored critic also degrades final performance, with the largest drop on the harder Latch subtask, where more frequent interventions make credit assignment across human-correction boundaries more critical.
The adaptation curves on Latch further show that the intervention-censored critic improves early data efficiency, yielding faster gains in both success rate and throughput.
Overall, the full model achieves the best final performance and the fastest online adaptation, showing that both wrench-context conditioning and intervention-censored value learning are important for stable, data-efficient online contact 
adaptation.

\FloatBarrier


\section{Conclusion}
\label{sec:conclusion}

We presented TORL-VLA, a VLA reinforcement learning framework for online contact adaptation with tactile-derived wrench feedback. TORL-VLA uses a wrench-aware VLA as an action-and-wrench reference model and refines VLA reference actions at contact-rich bottlenecks through a lightweight online actor-critic module.
To learn from human-intervened real-robot data, we introduce an intervention-censored critic that reduces credit-assignment bias by preventing post-intervention success from being over-credited to preceding policy-generated actions.
Real-robot experiments on long-horizon contact-rich manipulation tasks show that TORL-VLA achieves the best performance in success rates at both subtask and full-task levels, together with time-bounded execution efficiency.
Further ablation studies show that wrench-aware reference generation, online actor-critic refinement, and intervention-censored value learning all contribute to the observed gains.

\section{Limitations}
\label{sec:limitations}



Our evaluation is limited to latch-box manipulation with a few contact-rich bottlenecks, and broader validation across more objects, contact modes, environments, and robot embodiments is needed. TORL-VLA relies on reliable fingertip wrench measurements, but sensor noise, calibration errors, coordinate-frame misalignment, or hardware drift may degrade both measured feedback and predicted wrench references. Finally, intervention-based learning requires human-intervened data and assumes that human interventions can reliably indicate when autonomous execution is about to fail.




\bibliography{example}


\clearpage

\appendix
\section{Appendix}
\raggedbottom
\setlength{\textfloatsep}{10pt plus 2pt minus 2pt}
\setlength{\floatsep}{8pt plus 2pt minus 2pt}
\setlength{\intextsep}{8pt plus 2pt minus 2pt}

This appendix provides implementation and experimental details omitted from the main text due to space constraints. 
Appendix A.1 describes the latch-box manipulation benchmarks, the success and failure criteria, and the evaluation protocol used for autonomous evaluation and online adaptation. 
Appendix A.2 summarizes the robot system, tactile-to-wrench processing, demonstrations, and online replay. 
Appendix A.3 details the baseline adaptation and the implementation of the wrench-aware reference model and stage-specific online actor-critic refiners. 
Appendix A.4 presents additional diagnostics complementing the main results, including wrench and contact-state analysis and intervention-censored critic diagnostics.
Appendix A.5 reports deployment inference latency.
All task names, method names, and ablation names follow the notation in the main paper.

\subsection{Task and Evaluation Protocol}
\label{app:task_eval}

\subsubsection{Latch-Box Manipulation Tasks}

The real-robot evaluation is conducted on a latch-box manipulation benchmark with three contact-rich subtasks: \textbf{Coffee Cup}, \textbf{Latch}, and \textbf{Egg}.
Each subtask contains one or more local contact-rich bottleneck stages, where the stage estimator can activate the corresponding stage-specific online refiner during execution.
The workspace structure and subtask order are shared across methods, and all trials follow the same task-specific reset protocol, with object initial poses allowed to vary within predefined start regions.
The three subtasks are selected to cover different contact bottlenecks.
Coffee Cup evaluates tight insertion into a narrow holder under partial occlusion.
Latch evaluates contact-dependent mechanical engagement, where the robot must grip, flip, and lock a latch.
Egg evaluates fragile-object grasping and placement, where both insufficient and excessive contact can cause failure.

For full-task evaluation, the robot starts from the same latch-box initial configuration and executes the task sequence in one autonomous rollout.
A full-task rollout is counted as successful only if the robot completes Coffee Cup, Latch, and Egg sequentially without human intervention, environment reset, safety stop, object drop, or visible object damage.

\subsubsection{Success and Failure Criteria}

The subtasks are designed to expose contact states that may appear visually close to success while differing in physical interaction, such as partial insertion, incomplete latch engagement, or excessive grasping force.
All trials are reset according to the same task-specific reset protocol before evaluation.

The observable success and failure criteria are defined at the task-outcome level rather than by internal model signals or intermediate wrench traces, ensuring consistent evaluation across methods.
Table~\ref{tab:app_success_failure} summarizes these criteria for all subtasks and the full-task evaluation.
If any listed failure event occurs, the trial is marked as a failure even if human intervention could have recovered the task afterward.

\begin{table}[h]
\centering
\caption{\textbf{Observable success and failure criteria for autonomous evaluation.}}
\label{tab:app_success_failure}
\footnotesize
\setlength{\tabcolsep}{4.2pt}
\renewcommand{\arraystretch}{1.14}
\begin{tabularx}{\linewidth}{@{}p{0.15\linewidth}p{0.39\linewidth}p{0.39\linewidth}@{}}
\toprule
Task & Success condition & Failure condition \\
\midrule
Coffee Cup &
Pick cup; insert into holder; release; cup remains upright and stable. &
Holder miss; rim hang-up; partial insertion; large tilt; drop; unsafe contact. \\
\arrayrulecolor{gray!35}\midrule\arrayrulecolor{black}

Latch &
Grip latch; flip toward locking direction; press into locked state; latch remains secured after release. &
Missed latch; slip during flipping; missed locking edge; incomplete lock; rebound; unsafe force. \\
\arrayrulecolor{gray!35}\midrule\arrayrulecolor{black}

Egg &
Grasp egg; transfer to holder; release; egg remains stable without visible damage. &
Slip; drop; holder collision; rolling out after release; excessive grasping force; visible damage. \\
\arrayrulecolor{gray!35}\midrule\arrayrulecolor{black}

Full Task &
Complete Coffee Cup, Latch, and Egg sequentially in one autonomous rollout. &
Any subtask failure; human intervention; reset; safety stop; object drop; object damage. \\
\bottomrule
\end{tabularx}
\end{table}

\subsubsection{Evaluation Metrics and Protocol}

For both subtask and full-task success rates, each method is evaluated over 30 autonomous trials.
The same task objects, workspace structure, reset procedure, action representation, control frequency, and receding-horizon execution protocol are used for all methods.
During final autonomous evaluation, no human intervention, environment reset, safety-stop recovery, or online parameter update is allowed.
Online parameter updates are performed only during the online data-collection phase and are not performed while reporting the final success numbers.
For reference-model ablations, variants are evaluated directly on the three contact-rich subtasks.
For online-adaptation ablations, all variants use the same frozen wrench-aware reference model, checkpoint schedule, and matched online adaptation duration and interaction budget.

\subsection{System, Sensing, and Data}
\label{app:system_data}

\subsubsection{Robot and Sensor Setup}

The robot platform consists of a Piper robotic arm equipped with a Pika gripper from Agile Robots.
The visual input consists of three camera views: a global view captured by an Intel RealSense D435i camera, a fisheye view captured by the camera integrated in the Pika gripper, and a wrist view captured by an Intel RealSense D405 camera mounted on the Pika gripper.
The proprioceptive state includes the robot joint positions and the gripper state.
Tactile sensing is provided by two Tacta fingertip tactile sensors from EnsuringTech, mounted on the inner contact surfaces of the gripper fingers.
Each sensor has a compact size of 21 mm $\times$ 18.5 mm $\times$ 4.5 mm, and each tactile pad outputs a raw $6\times 8$ tactile array.

Policy observations are queried and logged at 20 Hz.
All modalities are timestamp-aligned during data collection.
For higher-rate tactile and proprioceptive streams, the latest measurement before the query timestamp is used.
For visual streams, the most recent synchronized image triplet is used.
All methods that use tactile-derived wrench feedback share the same tactile preprocessing, normalization statistics, and temporal window.

\subsubsection{Tactile-to-Wrench Processing}
\label{app:tactile_to_wrench}

The main paper defines the tactile-derived wrench observation as
\(\mathbf{w}_t=[\mathbf{w}_t^L,\mathbf{w}_t^R]\in\mathbb{R}^{12}\).
Here we clarify how this signal is obtained from the tactile sensor interface and how temporal wrench sequences are represented in implementation.
The vendor SDK exposes two levels of tactile readout.
The low-level interface returns \(48\) tactile points arranged on a \(6\times 8\) pad, with a 3D force vector at each point.
The high-level SDK interface provides a calibrated 6-DOF fingertip wrench estimate
\([f_x, f_y, f_z, \tau_x, \tau_y, \tau_z]\).
We denote this SDK-estimated fingertip wrench as
\(\tilde{\mathbf{w}}_t^i\in\mathbb{R}^{6}\) for each finger \(i\in\{L,R\}\).
The left and right fingertip outputs are concatenated in a fixed order to form the 12D wrench
\(\tilde{\mathbf{w}}_t=[\tilde{\mathbf{w}}_t^L,\tilde{\mathbf{w}}_t^R]\in\mathbb{R}^{12}\).

This vendor-provided wrench abstraction decouples the policy input from the raw tactile array layout.
It also provides a compact force--torque representation that serves as a unified downstream interface, rather than requiring the policy to process sensor-specific raw tactile layouts.

For the current measured wrench used in the online refinement context, we use the SDK-estimated 12D wrench, i.e., \(\mathbf{w}_t=\tilde{\mathbf{w}}_t\).
For temporal contact reasoning, the reference model uses \(J=10\) wrench samples over the most recent \(2\) seconds, corresponding to the recent wrench sequence \(\mathbf{W}_t=\{\mathbf{w}_{t-J+1},\ldots,\mathbf{w}_t\}\) in the main paper.
In implementation, before feeding this sequence into the wrench-history encoder, we apply a current-wrench-centered transform to the SDK-estimated wrench samples.
Specifically, the encoder input is formed as
\(\{\tilde{\mathbf{w}}_{t-J+1}-\tilde{\mathbf{w}}_t,\ldots,\tilde{\mathbf{w}}_{t-1}-\tilde{\mathbf{w}}_t,\mathbf{0}\}\).
Similarly, the future wrench target used in joint action--wrench prediction is represented by residuals
\(\tilde{\mathbf{w}}_{t+k}-\tilde{\mathbf{w}}_t\).
This implementation keeps the temporal structure described in the main paper while representing wrench sequences by relative wrench changes around the current contact state.

The same tactile-to-wrench preprocessing is used for all methods that consume wrench feedback, so differences between baselines and TORL-VLA come from the policy architecture and online adaptation mechanism rather than from different tactile preprocessing.

\subsubsection{Training Data for Reference Model and Online Adaptation}

Table~\ref{tab:app_training_data_budget} summarizes the main real-robot data budget used for the wrench-aware reference model and online actor-critic refinement.
Full-task demonstrations are used to train the wrench-aware VLA reference model, which predicts reference action chunks and future wrench sequences.
Main online interaction data are collected around contact-rich bottleneck stages for stage-specific online adaptation.
Warmup settings are reported in Table~\ref{tab:app_hyperparams}.

\begin{table}[h]
\centering
\caption{\textbf{Real-robot data budget for reference-model training and online adaptation.}
Main online interaction time denotes accumulated robot execution time during online data collection,
excluding human reset, object repositioning, and environment preparation.
The Egg subtask contains two contact-rich bottleneck stages: grasping and placement.}
\label{tab:app_training_data_budget}
\footnotesize
\setlength{\tabcolsep}{5pt}
\renewcommand{\arraystretch}{1.08}
\begin{tabular*}{0.88\linewidth}{@{\extracolsep{\fill}}lll@{}}
\toprule
Data source & Stage / Scope & Scale \\
\midrule
Demonstrations
& Full task
& 539 trajectories \\
\midrule
Main online interaction
& Coffee Cup
& 23.5 min \\
& Latch
& 40.3 min \\
& Egg: grasp
& 29.2 min \\
& Egg: place
& 27.8 min \\
\bottomrule
\end{tabular*}
\end{table}

\subsection{Implementation Details}
\label{app:implementation_details}

\subsubsection{Baseline Adaptation}
\label{app:baseline_adaptation}

We compare all methods under the same robot platform, action representation, training demonstrations, execution protocol, and evaluation procedure. We do not compare against numbers reported in prior work, since robot embodiments, tactile sensors, task definitions, and data distributions differ across papers. Instead, we adapt the baselines to the same \(\pi_{0.5}\)  backbone and, where applicable, the same dual-fingertip tactile-derived wrench interface used by TORL-VLA.

The \(\pi_{0.5}\)  baseline directly executes the pretrained VLA policy without tactile-derived wrench input. TA-VLA and ForceVLA are reimplemented on the same \(\pi_{0.5}\)  backbone and adapted to the dual-fingertip wrench observation. This keeps the comparison focused on whether tactile- or force-aware offline VLA designs can improve contact-rich manipulation under the same hardware and data setting. TORL-VLA w/o RL directly executes the learned wrench-aware reference model, without online RL refinement. RLT uses reference-guided online refinement, but it does not incorporate wrench input during online refinement and does not include the intervention-censored critic. TORL-VLA uses the wrench-aware reference model together with wrench-conditioned online refinement and intervention-censored value learning.

All online-adaptation variants use the same frozen wrench-aware reference model and the same online adaptation duration, while each variant collects its own online robot data. This ensures that the comparison between RLT, TORL-VLA, and online ablations focuses on the effect of wrench-context conditioning and intervention-censored value learning under a matched online interaction budget.

\subsubsection{Wrench-Aware Reference Model}

\begin{figure}[!t]
\vspace{-4pt}
\centering
\includegraphics[width=0.95\linewidth]{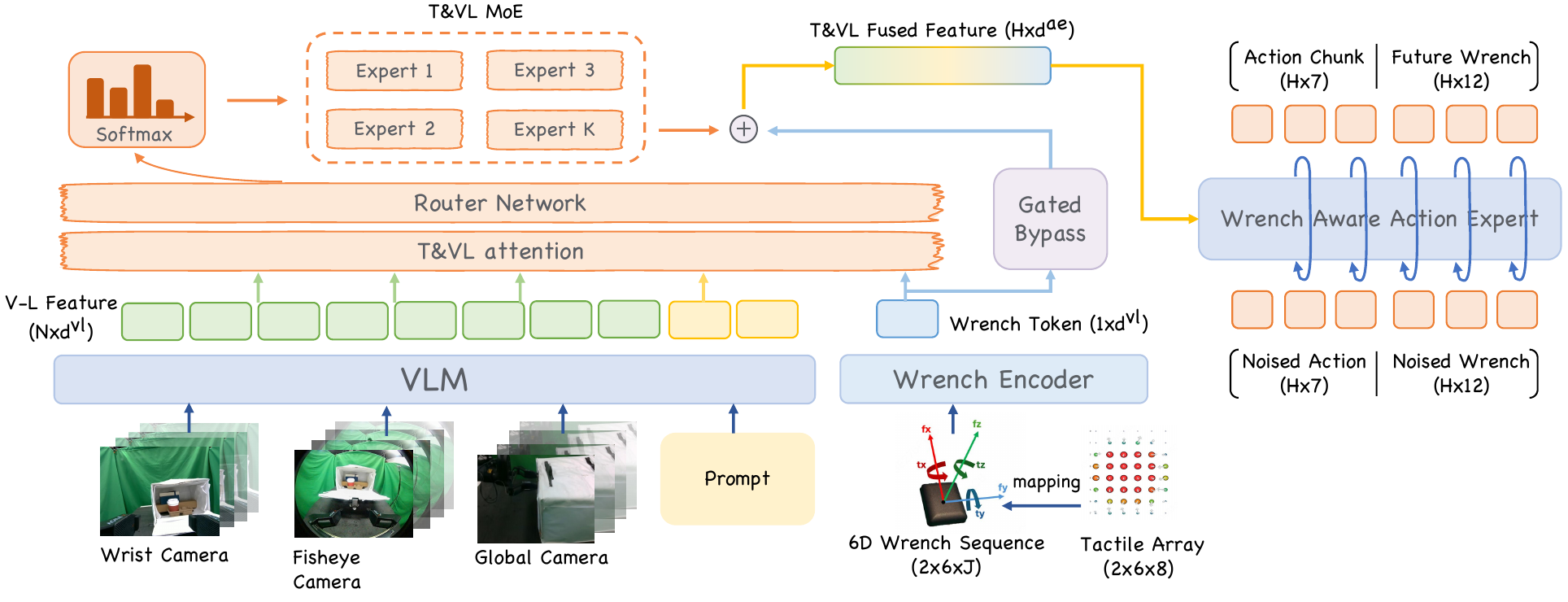}
\vspace{-2pt}
\caption{
\textbf{Implementation details of the wrench-aware VLA reference model.}
}
\label{fig:app_wrench_aware_reference_model}
\vspace{4pt}
\end{figure}

This subsection specifies the implementation details of Stage I, which are omitted from the main text for space.
At each timestep \(t\), the wrench-aware VLA reference model takes as input three camera views (wrist, fisheye, and global), a language instruction, robot proprioception, and tactile readings from two piezoelectric pads mounted on the inner surfaces of the gripper fingers.
As described in Appendix~\ref{app:tactile_to_wrench}, the tactile readings are converted into the current measured wrench \(\mathbf{w}_t\), and the recent wrench window is represented in a current-wrench-centered form before being fed into the wrench-history encoder.
The resulting history sequence is compressed into a fixed-size wrench token
\(\mathbf{z}_t^{\mathrm{tac}}\in\mathbb{R}^{1\times d_{\mathrm{vl}}}\)
by a lightweight MLP encoder.
In parallel, the three camera images, language instruction, and proprioception are encoded by the VLM backbone into visual-language prefix hidden states
\(\mathbf{P}_t\in\mathbb{R}^{N_p\times d_{\mathrm{vl}}}\).

Rather than injecting the wrench token early into the VLM, we adopt a late fusion strategy.
The wrench token is appended to \(\mathbf{P}_t\) after visual-language encoding, and a self-attention layer is applied to the concatenated sequence
\([\mathbf{P}_t;\mathbf{z}_t^{\mathrm{tac}}]\)
to produce the fused multimodal representation
\(\mathbf{H}_t\in\mathbb{R}^{(N_p+1)\times d_{\mathrm{vl}}}\).
This preserves the pretrained visual-semantic features while enabling cross-modal interaction with tactile-derived wrench feedback.
The fused representation is then passed through a soft-routing Mixture-of-Experts module, where a router network computes per-expert weights
\(\boldsymbol{\rho}_t=\mathrm{Softmax}(R_{\mathrm{tac}}(\mathbf{H}_t))\),
and the routed feature is obtained as
\(\mathbf{F}_t=\sum_{m=1}^{M}\boldsymbol{\rho}_{t,m}\odot E_m(\mathbf{H}_t)\).
This allows the model to dynamically route features according to the current contact phase.

To retain fine-grained physical information that may be diluted by the MoE transformations, we add a zero-initialized wrench-token bypass.
The routed feature and the original wrench token are projected to the action horizon and combined as
\(\mathbf{G}_t=\Pi_{\mathrm{tac}}(\mathbf{F}_t)+\alpha P_{\mathrm{tac}}^H(\mathbf{z}_t^{\mathrm{tac}})\in\mathbb{R}^{H\times d_{\mathrm{ae}}}\),
where \(\Pi_{\mathrm{tac}}\) maps the routed feature to the action horizon, \(P_{\mathrm{tac}}^H\) projects and expands the wrench token to horizon-aligned guidance tokens, and \(\alpha\) is a learnable scalar initialized to zero.
Thus, the bypass contributes nothing at the start of training and grows adaptively as the model learns to use wrench feedback.

Within the flow-matching action expert, the noisy joint generation variable
\(\mathbf{x}_{t,s}^{\mathrm{fm}}\in\mathbb{R}^{H\times(d_a+12)}\)
is processed at interpolation time \(s\in[0,1]\) to yield action-side hidden states \(\mathbf{S}_{t,s}\).
The wrench-conditioned guidance signal is added as a residual, producing
\(\tilde{\mathbf{S}}_{t,s}=\mathbf{S}_{t,s}+\mathbf{G}_t\).
A joint output head then predicts a single velocity field over the concatenated action--wrench target, which is split into action and wrench dimensions as
\(\hat{\mathbf{u}}_{t,s}=[\hat{\mathbf{u}}_{t,s}^{a},\hat{\mathbf{u}}_{t,s}^{w}]\).
Solving the learned flow from \(s=0\) to \(s=1\) yields the joint prediction: a reference action chunk
\(\hat{\mathbf{A}}_{t:t+H-1}\in\mathbb{R}^{H\times d_a}\)
that serves as the motion prior for the online actor, and a future wrench sequence
\(\hat{\mathbf{W}}_{t:t+H-1}\in\mathbb{R}^{H\times 12}\)
that provides the expected contact cue for wrench-conditioned action refinement in Stage II.
Following the preprocessing described in Appendix~\ref{app:tactile_to_wrench}, the future wrench target used in training is represented by current-centered residuals, i.e., \(\tilde{\mathbf{w}}_{t+k}-\tilde{\mathbf{w}}_t\).

\subsubsection{Online Reference Adaptation}

This subsection adds implementation details that are not fully specified in the main text.
During Stage II adaptation, the wrench-aware VLA is kept frozen and only the lightweight stage-specific actor-critic refiners are updated.

\noindent\textbf{Reference interface.}
The online refiner uses the same executable reference interface and context definition as in the main text.
In implementation, the frozen wrench-aware VLA predicts a 50-step action sequence and a 50-step future wrench sequence, and the first $K=10$ steps are used for online refinement and execution.
At 20 Hz, this corresponds to a 0.5 s replanning interval.

To obtain the compact VLA token $z_t^{\mathrm{vla}}$, we append a learned special token to the final hidden-state sequence of the frozen wrench-aware VLA and pass the augmented sequence through a lightweight token encoder.
The output at the special-token position is used as $z_t^{\mathrm{vla}}$.
The token encoder is trained before online adaptation and is frozen during online actor-critic updates.

\begin{wrapfigure}{r}{0.43\textwidth}
    \vspace{-6pt}
    \centering
    \includegraphics[width=0.41\textwidth]{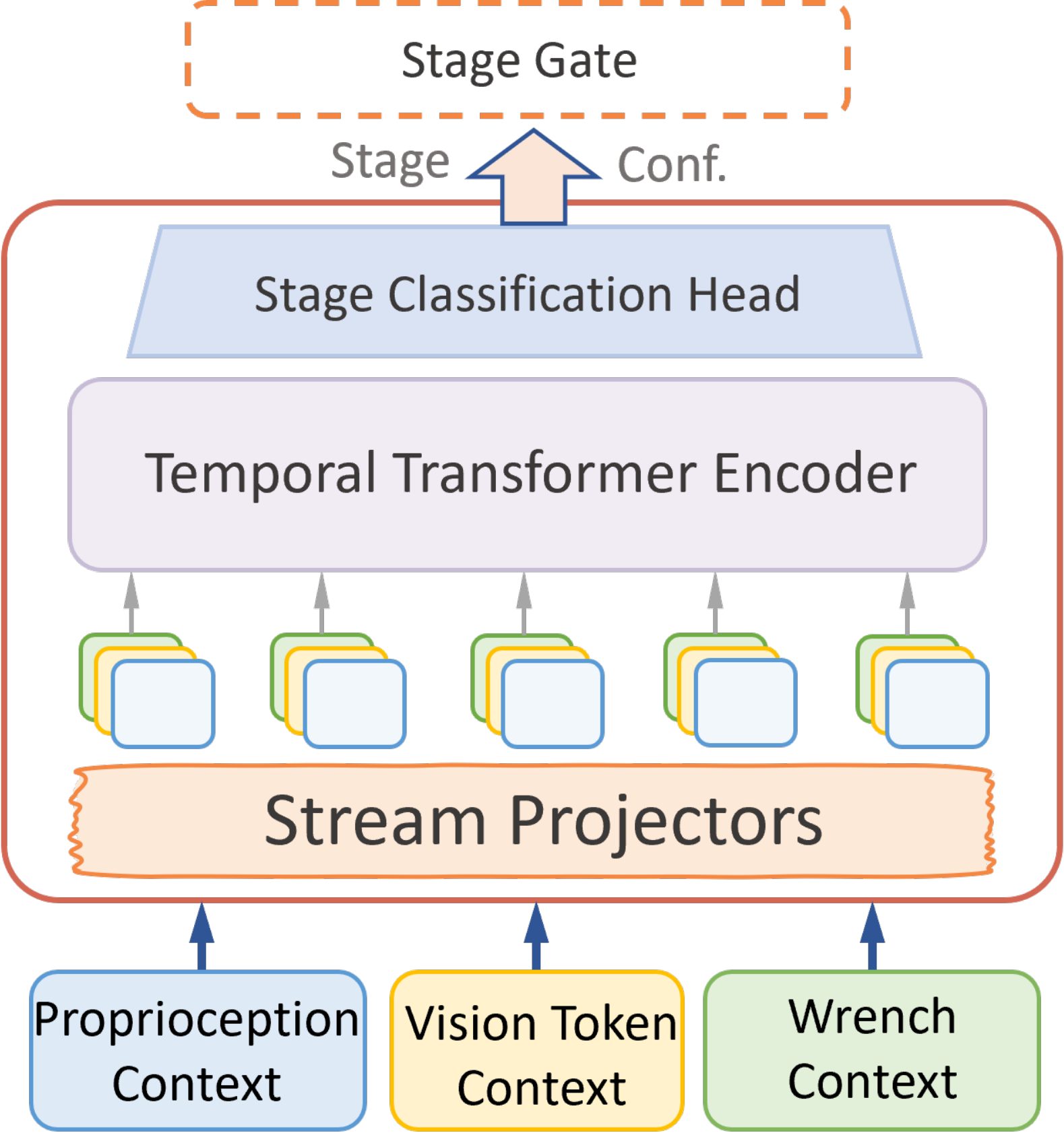}
    \caption{\textbf{Stage estimator architecture.}
    Policy-side features are projected, temporally aggregated, and used to predict a stage label and confidence for route filtering.}
    \label{fig:app_stage_estimator_architecture}
    \vspace{-6pt}
\end{wrapfigure}

\noindent\textbf{Stage estimator routing.}
As shown in Figure~\ref{fig:app_stage_estimator_architecture}, the stage estimator is a lightweight chunk-level prediction module rather than an action policy.
Its purpose is to decide how each VLA reference chunk should be executed during Stage II adaptation.
At each VLA query, the frozen wrench-aware VLA first produces the reference action chunk and the policy-side features used by the online module.
The stage estimator reuses these features, including the compact VLA token context, proprioception context, and wrench context, to predict a discrete stage label and its confidence.
The predicted stage label is then mapped to a runtime route before executing the next $K$-step action chunk.

The route mapping separates full-task execution into base execution and local contact-window refinement.
Labels outside predefined contact windows are mapped to the base route, where the frozen VLA reference chunk is executed directly.
Only labels corresponding to contact-critical windows activate the associated stage-specific actor-critic refiner.
Therefore, the stage estimator does not assign an entire subtask to an online refiner.
Instead, it only selects short windows in which contact-aware refinement is needed.
For example, approach and inter-stage transport are usually handled by the frozen VLA, while online refinement is activated around cup insertion, latch engagement/locking, egg grasping, and egg placement.

To avoid unstable switching caused by isolated prediction noise, we apply a simple route gate before execution.
A predicted route is accepted only when the stage confidence exceeds $0.9$ and the same mapped route is predicted for two consecutive chunks.
Otherwise, the system keeps the previous stable route or falls back to the base route.
Because the stage estimator reuses features already produced at the VLA query, this routing step does not require an additional VLA forward pass.

\begin{figure}[t]
\centering
\includegraphics[width=\linewidth]{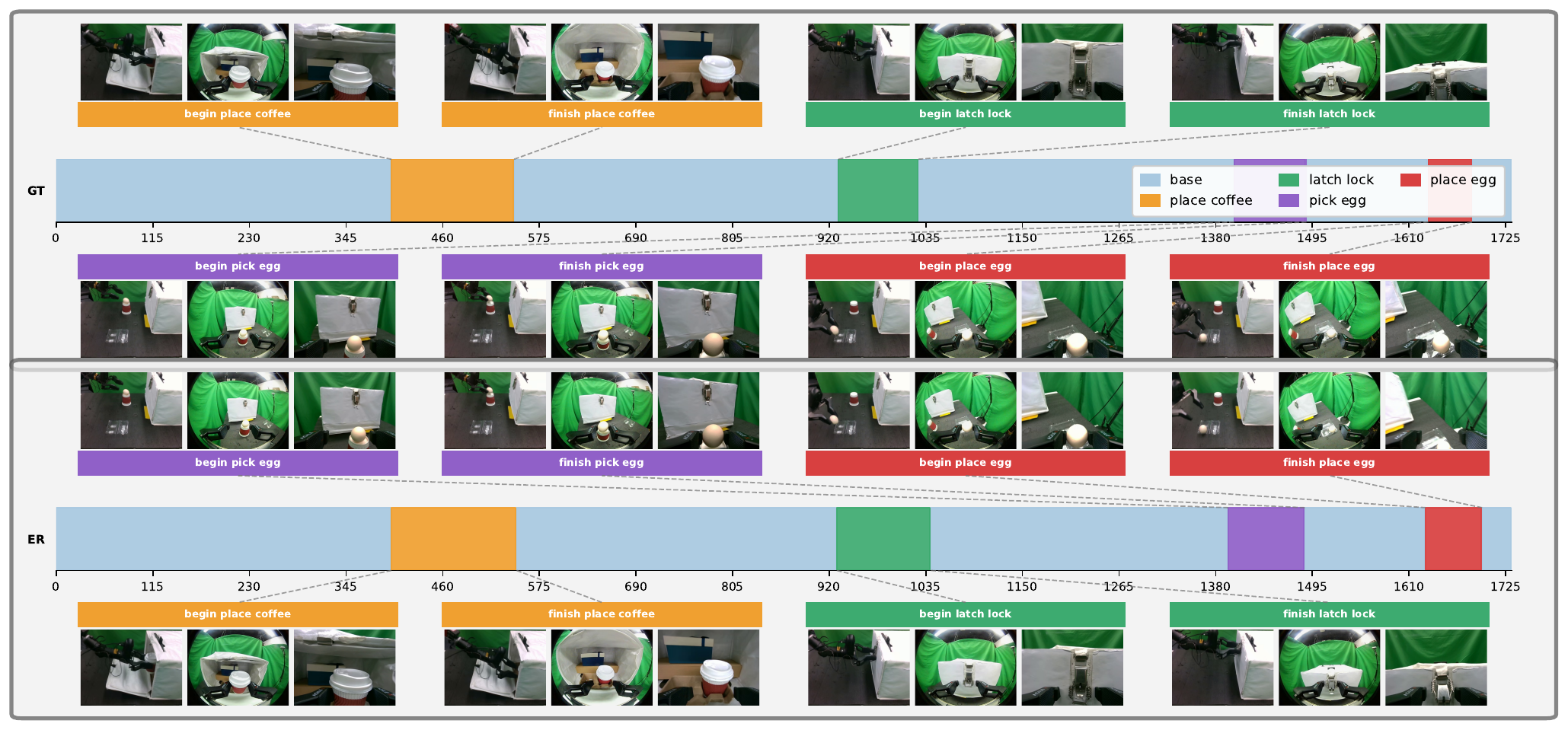}
\caption{\textbf{Detailed alignment of predicted stages against ground truth over full-task execution.}
The visualization trace shows chunk-level stage estimates aligned with sequential camera observations.
Minor discrepancies mainly appear at continuous phase boundaries, while the physical manipulation states in GT and ER remain consistent across stable intervals.}
\label{fig:app_stage_estimator_prediction}
\end{figure}

Figure~\ref{fig:app_stage_estimator_prediction} compares the estimated stages with ground-truth annotations over a full-task rollout.
The stage estimator achieves an absolute semantic accuracy of 95.5\%.
The remaining mismatches are concentrated near the continuous boundaries between adjacent stages.
This is expected because the physical transition between two neighboring phases is gradual rather than instantaneous, and manual stage annotations at such boundaries can vary slightly.
Direct inspection of the aligned camera observations shows that the physical manipulation states corresponding to the ground truth (GT) and estimation results (ER) are visually and functionally similar across stable intervals.
Thus, these boundary-level discrepancies mainly reflect annotation ambiguity around phase transitions rather than incorrect recognition of stable task stages.
In deployment, the confidence and temporal-stability gate further reduces the chance that such boundary fluctuations lead to unstable actor switching.
Moreover, the stage-specific online actors operate on continuous wrench-conditioned contexts, making the refinement robust to small timing shifts around the beginning or end of a contact window.

When a chunk is routed to a contact window, the corresponding stage-specific actor-critic refiner is used.
The actor embeds the input groups in $c_t$, concatenates the features, and directly outputs the refined executable chunk:
\begin{equation}
    A_t^\phi = \pi_{\phi_{g_t}}(c_t).
\end{equation}
The actor does not predict an explicit residual added to $A_t^{\mathrm{ref}}$.
Action chunks are normalized internally for training and converted back to executable robot commands before deployment.

\noindent\textbf{Replay buffer and control-source labels.}
Before main online adaptation, each stage-specific refiner is initialized with warmup data, where execution mainly follows the VLA reference and human intervention is applied near failure or unsafe contact.
After warmup, the corresponding actor snapshot is used in closed-loop execution.

Online data are stored at the executed $K$-step chunk level.
Each replay entry contains $A_t^{\mathrm{ref}}$, the executed chunk $A_t^{\mathrm{data}}$, robot proprioception, the current measured wrench, recent measured wrench history, the aligned predicted wrench cue $\hat{W}_t^{\mathrm{ref}}$, rewards, terminal flag, next context, and per-step control-source labels.
For each low-level step $j$, $\sigma_{t,j}\in\{\mathrm{policy},\mathrm{human}\}$ records whether the action was generated by the learned policy or by human intervention.
These labels are not actor observations; they are used only for behavior regularization, intervention-boundary target construction, and diagnostics.

\FloatBarrier

\subsubsection{Training Details and Hyperparameters}
\label{app:training_details}

Table~\ref{tab:app_hyperparams} summarizes the key hyperparameters used for reference generation and online adaptation.
We report the settings that define the reference interface, execution schedule, online update cadence, intervention-boundary construction, and stage-specific actor-critic refiners.
Since the Egg subtask contains two contact-rich bottleneck stages, we use separate actor-critic refiners for egg grasping and egg placement.

\begin{table}[!htbp]
\centering
\caption{\textbf{Key hyperparameters for reference generation and online adaptation.}}
\label{tab:app_hyperparams}
\footnotesize
\setlength{\tabcolsep}{2.8pt}
\renewcommand{\arraystretch}{1.08}

\begin{minipage}[t]{0.46\linewidth}
\centering
\begin{tabularx}{\linewidth}{@{}X>{\centering\arraybackslash}p{0.30\linewidth}@{}}
\toprule
Reference / execution & Value \\
\midrule
Reference horizon $H$ & 50 \\
Executed horizon $K$ & 10 \\
Action / wrench dimension & 7 / 12 \\
Robot command representation & delta chunk \\
Control frequency & 20 Hz \\
Replanning interval & 0.5 s \\
Wrench history length & 10 \\
Wrench history window & 2.0 s \\
Future wrench horizon & 50 \\
Future wrench loss $\lambda_w$ & 0.3 \\
VLA token dimension $z_t^{\mathrm{vla}}$ & 2048 \\
\bottomrule
\end{tabularx}
\end{minipage}
\hspace{0.035\linewidth}
\begin{minipage}[t]{0.46\linewidth}
\centering
\begin{tabularx}{\linewidth}{@{}X>{\centering\arraybackslash}p{0.30\linewidth}@{}}
\toprule
Online learning / AC & Value \\
\midrule
Discount factor $\gamma$ & 0.99 \\
Update-to-data ratio & 5 \\
Critic update frequency & every step \\
Actor update frequency & every 2 steps \\
Target-network EMA $\tau$ & 0.005 \\
Actor snapshot interval & 100 steps \\
Checkpoint interval & 1000 steps \\
Intervention cost $c_{\mathrm{int}}$ & 1.0 \\
Human-control threshold $\rho_{\mathrm{int}}$ & 0.5 \\
Warmup replay chunks & 300 \\
Warmup gradient updates & 5k \\
Hidden dimension & 256 \\
\midrule
Coffee Cup AC layers & 3 / 3 \\
Egg-grasp AC layers & 3 / 3 \\
Egg-place AC layers & 3 / 3 \\
Latch AC layers & 4 / 4 \\
\bottomrule
\end{tabularx}
\end{minipage}
\end{table}
\FloatBarrier

Online learning uses replay-buffer updates.
For each newly added replay transition, the learner performs five gradient updates on average; the critic is updated at every learner step, while the actor is updated every two learner steps.
Checkpoints are saved every 1000 learner steps.

These settings specify the TORL-VLA implementation used in our online adaptation experiments, including the execution horizon, control frequency, warmup schedule, online update cadence, and stage-specific actor-critic architecture.

\subsection{Additional Diagnostics}
\label{app:additional_analysis}

\subsubsection{Wrench and Contact-State Analysis}
\label{app:wrench_contact_analysis}

This section analyzes tactile-derived wrench signals during contact-rich execution. The goal is not to introduce an additional benchmark metric, but to show how measured wrench and predicted future wrench behave in the task phases visualized below.

\begin{figure*}[t]
\centering
\includegraphics[width=0.92\textwidth]{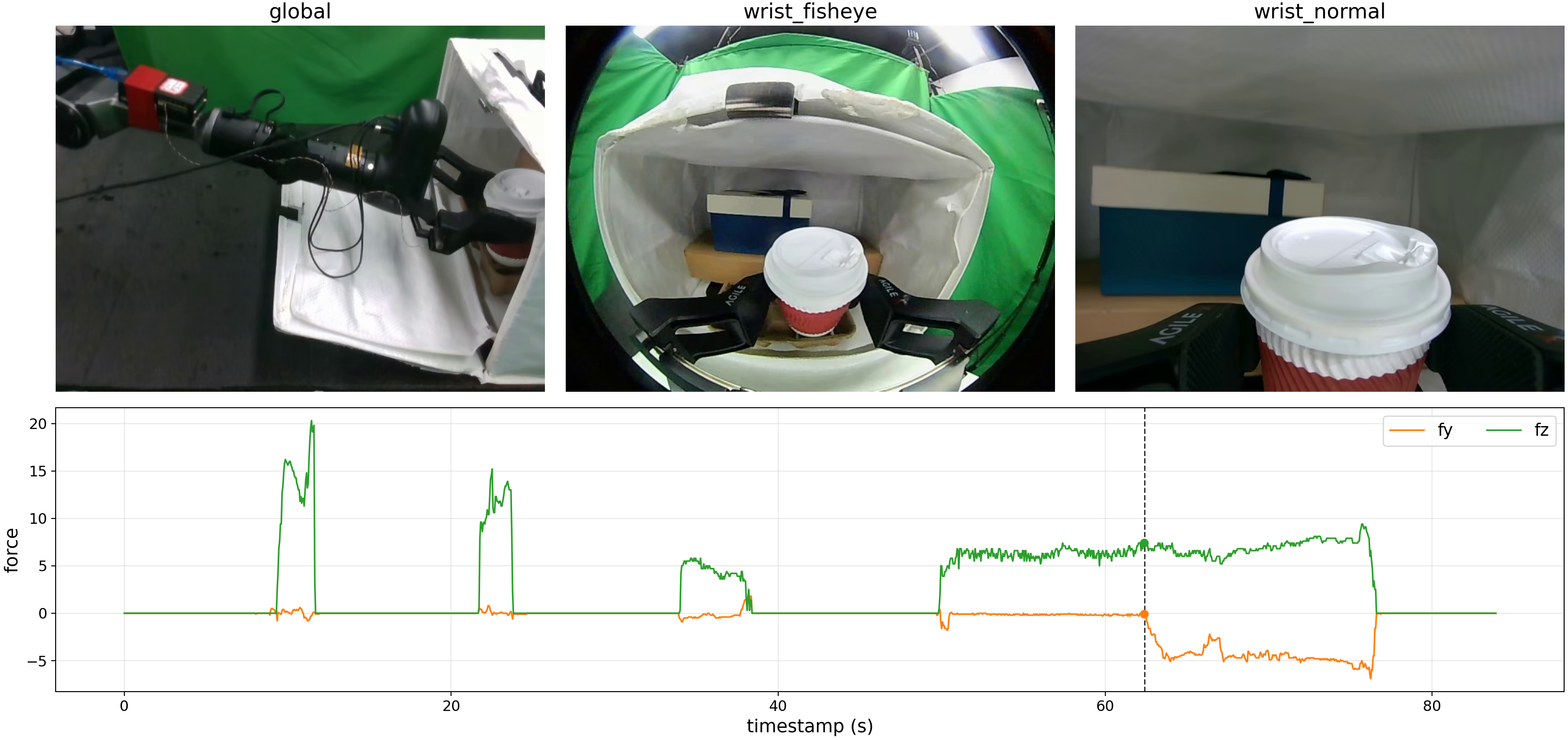}
\caption{
Measured contact evolution in the cup-insertion task. The wrench curves are measured from the right-fingertip tactile sensor. The gripper maintains the normal force $f_z$ to stably hold the cup while applying an appropriate downward force $f_y$ during insertion.
}
\label{fig:app_measured_wrench_cup}
\end{figure*}

\noindent\textbf{Measured contact evolution.}
Figures~\ref{fig:app_measured_wrench_latch} and~\ref{fig:app_measured_wrench_cup} show representative real-execution segments from the latch-locking and cup-insertion tasks. The wrench curves shown in this analysis are measured from the right-fingertip tactile sensor. The latch sequence in Figures~\ref{fig:app_measured_wrench_latch} covers the process from the initial latch grasp to latch flipping. During this process, the commanded action changes smoothly, while the measured fingertip wrench exhibits clear temporal variation. In the cup task shown in Figures~\ref{fig:app_measured_wrench_cup}, the gripper must maintain a normal force $f_z$ to stably hold the cup, while also applying an appropriate downward force $f_y$ during insertion to place the cup into the holder. These examples show that the contact state in our tasks is not a simple binary contact/no-contact variable, but evolves continuously with the interaction among the robot, the manipulated object, and the environment. Beyond characterizing normal contact evolution, the wrench signal also provides cues for identifying deviations from the expected execution. For example, grasping the latch too tightly may introduce excessive contact forces that prevent the latch from being flipped. Similarly, during cup insertion, a slight positional offset may be difficult to detect from visual observations alone, but it can be revealed by abnormal variations in \(f_y\).

\begin{figure*}[t]
\centering
\includegraphics[width=0.86\textwidth]{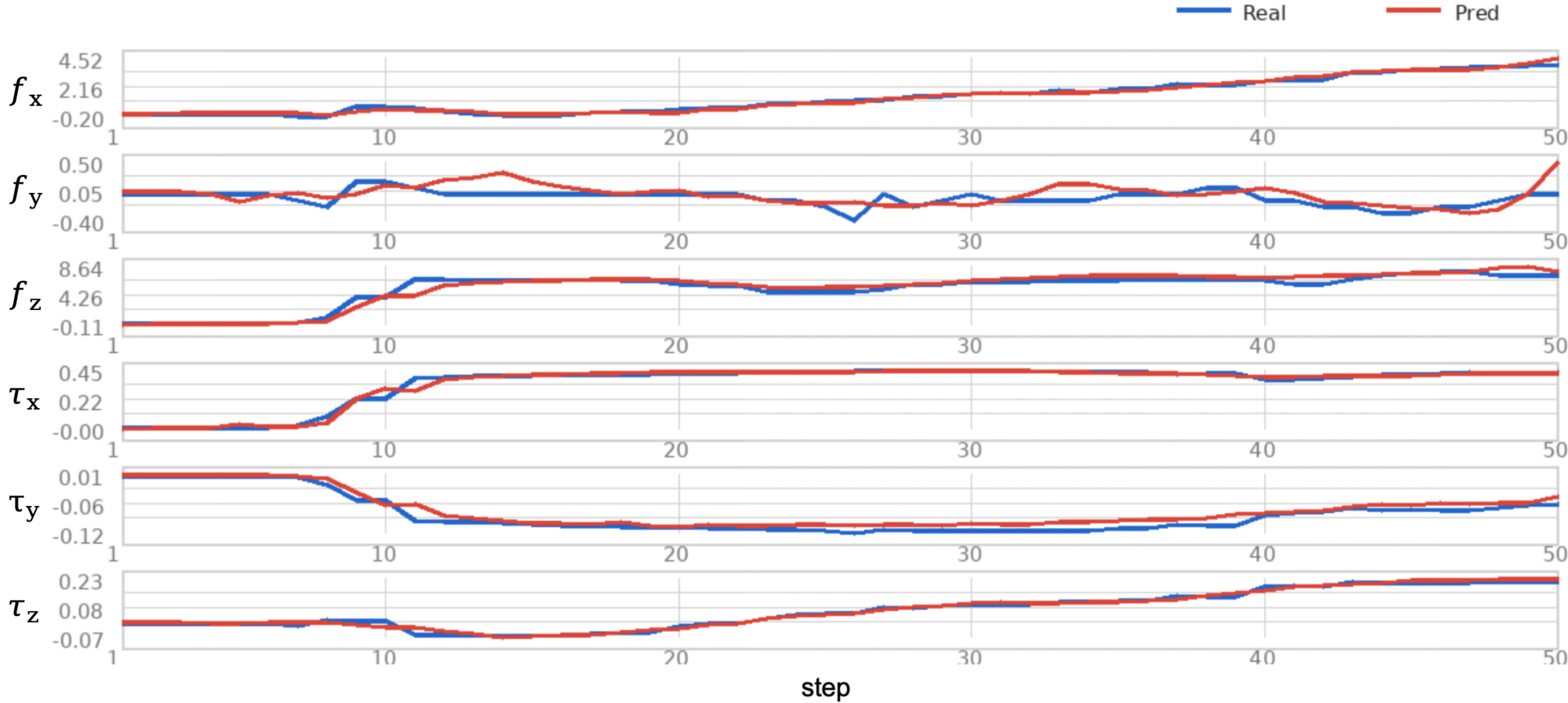}
\caption{
Detailed six-dimensional future-wrench prediction on a latch-locking segment. The figure compares the predicted and measured six-dimensional wrench trajectories for the left-fingertip tactile sensor, including the three force components and three torque components. This view complements the norm-based summaries in Figure~\ref{fig:app_future_wrench_examples_all} by showing whether the model captures the component-wise contact evolution.
}
\label{fig:app_future_wrench_6d_latch}
\end{figure*}

\noindent\textbf{Future-wrench prediction.}
Figure~\ref{fig:app_future_wrench_examples_all} evaluates whether the reference model can predict future contact trends. Each panel pairs the current visual observation and the observation after 50 action steps with the predicted and measured fingertip wrench norms. The examples cover latch locking, cup insertion, and egg grasping, thereby testing the prediction head under mechanism contact, environmental contact, and delicate object contact. These curves suggest that the reference model anticipates the contact consequences of its proposed future action.  Figure~\ref{fig:app_future_wrench_6d_latch} further expands one latch-locking segment into the left-fingertip components, showing the predicted and measured force--torque trajectories along all six dimensions.

\noindent\textbf{Beyond history extrapolation.}
A potential concern is that the future-wrench head may simply extrapolate the trend of the input wrench history. To probe this, we examine the examples in subpanels (a1), (b1), and (c1) of Figure~\ref{fig:app_future_wrench_examples_all}. In these cases, the input wrench histories are equal to zero before substantial contact is established. A history-copying predictor would therefore remain near zero or continue the previous flat trend. Instead, the model predicts a future rise or change in the fingertip wrench while generating the corresponding reference action. This suggests that the reference model infers upcoming contact dynamics from the visual state and the shared action--wrench generation context, rather than merely extrapolating short-term wrench history.
\FloatBarrier

\subsubsection{Intervention-Censored Critic Diagnostics}
\label{app:ic_critic_diagnostics}

This section provides process-level diagnostics for the intervention-censored critic on the Latch-Lock stage, i.e., the locking contact window within the Latch subtask.
We use this stage because it contains sufficient policy-to-human boundary contexts for stable analysis.
These diagnostics are not additional final evaluation metrics; they are used only to inspect whether $Q_{\mathrm{ic}}$ captures the intended intervention-risk signal and whether the learned actor moves away from actions with lower intervention-censored value.
Unless otherwise specified, all diagnostics are computed from the final Latch-Lock online replay of the full model, and critic values are evaluated using the final critic checkpoint.

\noindent\textbf{Boundary contexts and diagnostic scores.}
For diagnostics, we define a clean policy-generated chunk as a mostly policy-generated chunk that is not followed by an intervention-dominated chunk.
A policy-to-human boundary chunk is a mostly policy-generated chunk immediately preceding an intervention-dominated chunk, using the same threshold as the intervention-boundary mask in the main method.
A boundary context denotes the actor context $c_t$ associated with such a boundary chunk.
These diagnostic labels are used only for offline analysis and are not provided to the actor as observations.

All critic values are reported as the minimum over the double-Q pair.
For AUC-based boundary-risk ranking, we use $-Q(c,A)$ as the score: lower critic values indicate lower predicted value, and for $Q_{\mathrm{ic}}$ this is intended to correspond to higher intervention risk.
Thus, a larger AUC means that a critic more reliably ranks boundary chunks above clean policy-generated chunks under this diagnostic score.

\noindent\textbf{Boundary-context distribution and actor behavior.}
Table~\ref{tab:app_boundary_prevalence_actor} reports two diagnostics on the final Latch-Lock replay.
The left table shows how policy-to-human boundary contexts are distributed across online episodes, indicating that they are recurring contact-risk events rather than a few isolated outliers.
The right table compares the logged pre-intervention action, the VLA reference action, and the final actor action on the same 165 boundary contexts.
The final actor achieves higher average $Q_{\mathrm{ic}}$ value than both the logged action and the VLA reference, suggesting that the learned actor moves away from lower-IC-value actions on these boundary contexts.
We call a boundary context \emph{IC-hinge-active} when the reference-relative IC hinge in the actor loss is active, i.e.,
$\min_i Q_{\mathrm{ic}}^{(i)}(c,A^\phi) < \min_i Q_{\mathrm{ic}}^{(i)}(c,A^{\mathrm{ref}})$.


\begin{table}[!htbp]
\centering
\caption{\textbf{Boundary-context distribution and actor behavior.}
Left: episode-level distribution of policy-to-human boundary contexts in the final Latch-Lock replay.
Right: action comparison on the same 165 boundary contexts using the final IC critic.}
\label{tab:app_boundary_prevalence_actor}
\footnotesize
\setlength{\tabcolsep}{3.2pt}
\renewcommand{\arraystretch}{1.08}

\resizebox{0.98\linewidth}{!}{%
\begin{tabular}{@{}c@{\hspace{2.8em}}c@{}}
\begin{tabular}{@{}lc@{}}
\toprule
Metric & Value \\
\midrule
Episodes with replay & 162 \\
Boundary contexts & 165 \\
Episodes $\geq$ 1 boundary & 125 / 162 = 77.2\% \\
Episodes $\geq$ 2 boundaries & 32 / 162 = 19.8\% \\
Episodes $\geq$ 3 boundaries & 8 / 162 = 4.9\% \\
Mean per episode & 1.02 \\
Median per episode & 1 \\
90th percentile & 2 \\
Maximum per episode & 3 \\
\bottomrule
\end{tabular}
&
\begin{tabular}{@{}lc@{}}
\toprule
Actor diagnostic & Value \\
\midrule
Logged $Q_{\mathrm{ic}}(c,A^{\mathrm{data}})$ & -1.003 \\
Reference $Q_{\mathrm{ic}}(c,A^{\mathrm{ref}})$ & -0.573 \\
Final actor $Q_{\mathrm{ic}}(c,A^\phi)$ & -0.271 \\
$\Delta Q_{\mathrm{ic}}$: actor vs. reference & +0.301 \\
$\Delta Q_{\mathrm{ic}}$: actor vs. logged data & +0.731 \\
IC-hinge-active boundary contexts & 21 / 165 = 12.73\% \\
\bottomrule
\end{tabular}
\end{tabular}%
}
\end{table}

\noindent\textbf{Boundary-risk separation.}
Table~\ref{tab:app_ic_value_sep} compares $Q_{\mathrm{ic}}$ and $Q_{\mathrm{task}}$ on clean policy-generated chunks and policy-to-human boundary chunks.
Clean Mean and Boundary Mean denote the average critic values on the two groups, and Clean--Boundary is their difference.
AUC is computed using $-Q(c,A)$ as the boundary-risk score.
The IC critic separates boundary chunks from clean chunks much more clearly than the task critic, supporting the separation between task-return estimation and intervention-risk modeling.

\begin{table}[!htbp]
\centering
\caption{\textbf{Boundary-risk separation by task and IC critics.}
AUC is computed using $-Q(c,A)$ as the boundary-risk score; for $Q_{\mathrm{ic}}$, lower values are intended to indicate higher intervention risk.}
\label{tab:app_ic_value_sep}
\footnotesize
\setlength{\tabcolsep}{3.5pt}
\renewcommand{\arraystretch}{1.08}
\begin{tabular}{lcccc}
\toprule
Critic / Action
& Clean Mean
& Boundary Mean
& Clean--Boundary
& AUC $\uparrow$ \\
\midrule
$Q_{\mathrm{ic}}(c,A^{\mathrm{data}})$
& 0.073 & -1.003 & 1.076 & 0.999 \\
$Q_{\mathrm{ic}}(c,A^{\mathrm{ref}})$
& -0.007 & -0.573 & 0.565 & 0.913 \\
$Q_{\mathrm{ic}}(c,A^{\phi})$
& 0.091 & -0.271 & 0.363 & 0.806 \\
\midrule
$Q_{\mathrm{task}}(c,A^{\mathrm{data}})$
& 0.333 & 0.247 & 0.086 & 0.621 \\
$Q_{\mathrm{task}}(c,A^{\mathrm{ref}})$
& 0.316 & 0.241 & 0.074 & 0.626 \\
$Q_{\mathrm{task}}(c,A^{\phi})$
& 0.351 & 0.272 & 0.079 & 0.618 \\
\bottomrule
\end{tabular}
\end{table}

\noindent\textbf{Actor checkpoint comparison.}
Table~\ref{tab:app_actor_resolution_diag} compares actor checkpoints saved at learner global steps 10k, 20k, and 33k.
To avoid comparing values from drifting critic checkpoints, we use the final $Q_{\mathrm{ic}}$ as a fixed diagnostic scorer for all actor checkpoints.
The comparison is retrospective: all checkpoints are evaluated on the same final boundary-context set, although not all of these contexts were present in replay at earlier checkpoints.
A boundary context is counted as IC-hinge-active at a checkpoint if the actor action at that checkpoint has lower IC value than the VLA reference under the same context.
A 10k IC-hinge-active context is considered resolved if it is no longer IC-hinge-active under the final actor.
The number of IC-hinge-active boundary contexts decreases from $89/165$ at 10k to $21/165$ at the final 33k checkpoint, indicating that the actor progressively reduces boundary contexts where its action has lower IC value than the VLA reference.

\begin{table}[!htbp]
\centering
\caption{\textbf{Actor checkpoint comparison on boundary contexts.}
The final IC critic is used as a fixed scorer on the same final boundary-context set.
Checkpoints are indexed by learner global step.
Margin statistics are computed only on the 89 contexts that are IC-hinge-active at the 10k checkpoint.}
\label{tab:app_actor_resolution_diag}
\footnotesize
\setlength{\tabcolsep}{5pt}
\renewcommand{\arraystretch}{1.08}
\begin{tabular}{lc}
\toprule
Retrospective actor diagnostic & Value \\
\midrule
IC-hinge-active contexts at 10k global-step checkpoint & 89 / 165 = 53.94\% \\
IC-hinge-active contexts at 20k global-step checkpoint & 66 / 165 = 40.00\% \\
IC-hinge-active contexts at final 33k global-step checkpoint & 21 / 165 = 12.73\% \\
10k IC-hinge-active contexts resolved by final actor & 71 / 89 = 79.8\% \\
Mean margin on 10k IC-hinge-active contexts & -0.207 \\
Mean margin on the same contexts at final 33k checkpoint & +0.213 \\
Mean margin improvement on the same contexts & +0.421 \\
\bottomrule
\end{tabular}
\end{table}

\noindent\textbf{Online intervention behavior.}
Table~\ref{tab:app_online_intervention_diag} reports rollout-level process diagnostics during late online training.
These statistics are not used as final autonomous evaluation metrics.
Human-Controlled Step Ratio is computed at the low-level control-step level as the fraction of steps involving human control, and Avg. Human Interventions counts human takeover events per episode.
Compared with the variant without the IC critic under the same late-training window, the full model shows lower human-intervention reliance and more intervention-free episodes during late online training.
This is consistent with the intended role of $Q_{\mathrm{ic}}$: reducing policy behaviors that tend to lead into human correction.

\begin{table}[!htbp]
\centering
\caption{\textbf{Latch-Lock online intervention diagnostics.}
Statistics are computed over late online training rollouts and are not used as final autonomous evaluation metrics.
Human-Controlled Step Ratio is computed at the low-level control-step level, and Avg. Human Interventions counts human takeover events per episode.}
\label{tab:app_online_intervention_diag}
\footnotesize
\setlength{\tabcolsep}{5pt}
\renewcommand{\arraystretch}{1.12}
\begin{tabular*}{0.92\linewidth}{@{\extracolsep{\fill}}lccc@{}}
\toprule
Variant / Window
& \makecell[c]{Human-Controlled\\Step Ratio $\downarrow$}
& \makecell[c]{Avg. Human\\Interventions $\downarrow$}
& \makecell[c]{Intervention-Free\\Episodes $\uparrow$} \\
\midrule
w/o IC Critic, last 100 ep.
& 27.78\% & 13.40 & 11.0\% \\
Full model, last 100 ep.
& 25.32\% & 10.68 & 29.0\% \\
w/o IC Critic, last 50 ep.
& 25.42\% & 13.10 & 12.0\% \\
Full model, last 50 ep.
& 18.12\% & 7.48 & 40.0\% \\
\bottomrule
\end{tabular*}
\end{table}
\FloatBarrier

\subsection{Deployment Efficiency}
\label{app:deployment_latency}

\subsubsection{Deployment Inference Latency}

Table~\ref{tab:app_latency} reports deployment-time inference latency measured on a single NVIDIA RTX 4090 GPU with batch size one.
We measure online policy computation only, excluding robot communication, observation acquisition, low-level controller execution, and environment preparation.
All policy-query latencies are averaged over 30 timed warm-call queries after excluding the first JIT/initialization call and using five warmup queries.

During a TORL-VLA policy query, the frozen wrench-aware VLA generates the reference action sequence and predicted future wrench sequence.
The stage estimator performs chunk-level routing using policy-query features and does not require an additional VLA forward pass.
When the current chunk is routed to a contact-critical stage, the corresponding stage-specific online actor refines the reference action chunk.
The reported TORL-VLA total is measured end-to-end for this actor-refinement route, rather than obtained by summing separately measured module times.
The base $\pi_{0.5}$ row is included only as a reference-policy comparison and is not part of the TORL-VLA total.

\begin{table}[!htbp]
\centering
\caption{\textbf{Deployment inference latency.}
Mean warm-call latency is measured on a single NVIDIA RTX 4090 GPU with batch size one.}
\label{tab:app_latency}
\footnotesize
\setlength{\tabcolsep}{5pt}
\renewcommand{\arraystretch}{1.08}
\begin{tabular}{lcc}
\toprule
Module / Query & Timed Calls & Mean Latency \\
\midrule
Base $\pi_{0.5}$ policy query (comparison) & 30 & 94.7 ms \\
TORL-VLA wrench-aware reference query & 30 & 100.2 ms \\
Stage estimator routing & 30 & 1.7 ms \\
Stage-specific online actor & 30 & 1.1 ms \\
\midrule
End-to-end TORL-VLA query with actor refinement & 30 & 103.0 ms \\
\bottomrule
\end{tabular}
\end{table}

Compared with the base $\pi_{0.5}$ policy query, the full TORL-VLA actor-refinement route increases mean policy-query latency by 8.3 ms, corresponding to an 8.8\% relative overhead.
The stage estimator and online actor add only a few milliseconds of latency, indicating that TORL-VLA introduces lightweight deployment overhead while retaining the same real-time policy-query regime.

\begin{figure*}[t]
\centering
\setlength{\tabcolsep}{3pt}
\begin{tabular}{c}
\includegraphics[width=0.86\textwidth]{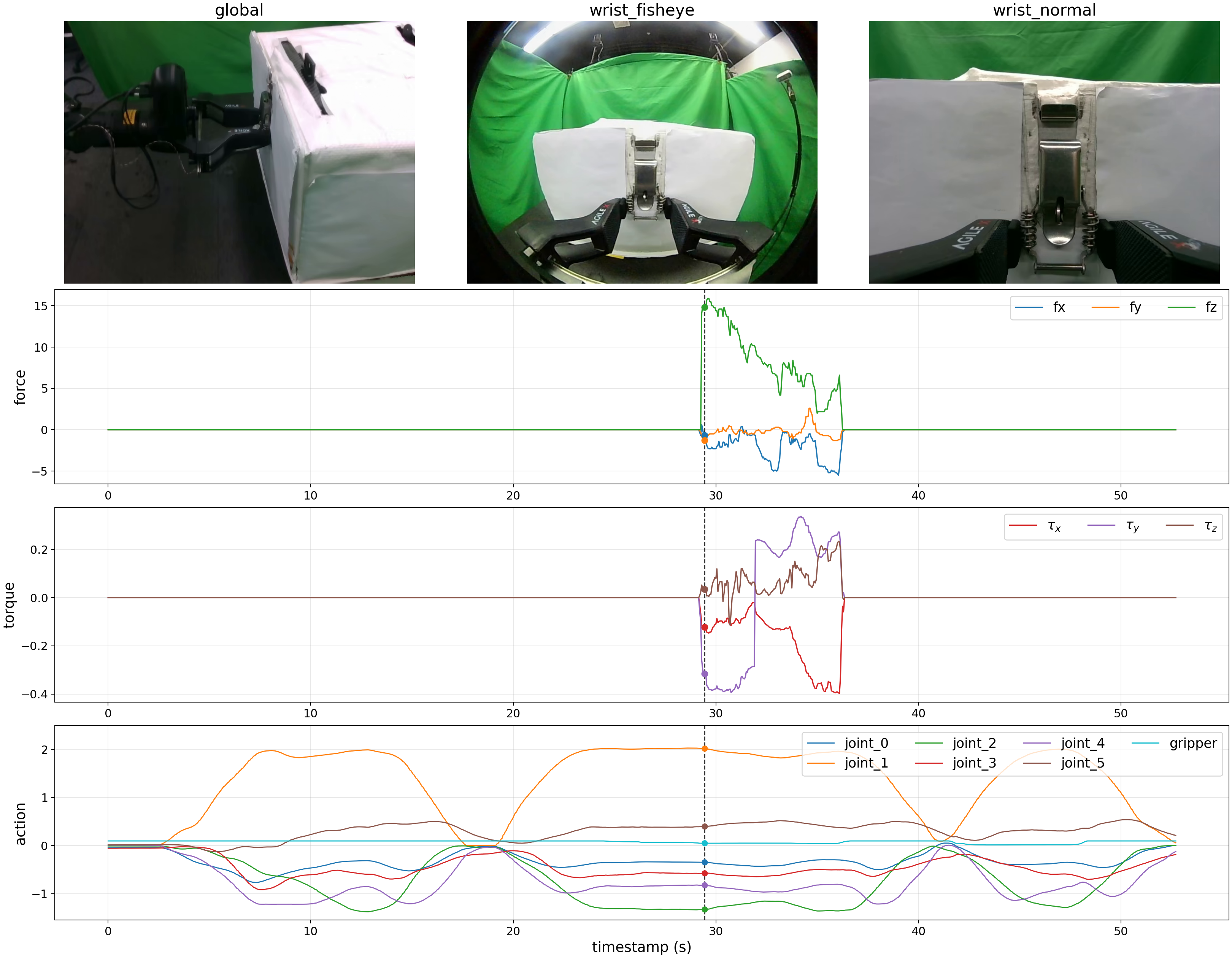} \\
\small (a) Initial latch-grasping phase \\
\addlinespace[0.6em]
\includegraphics[width=0.86\textwidth]{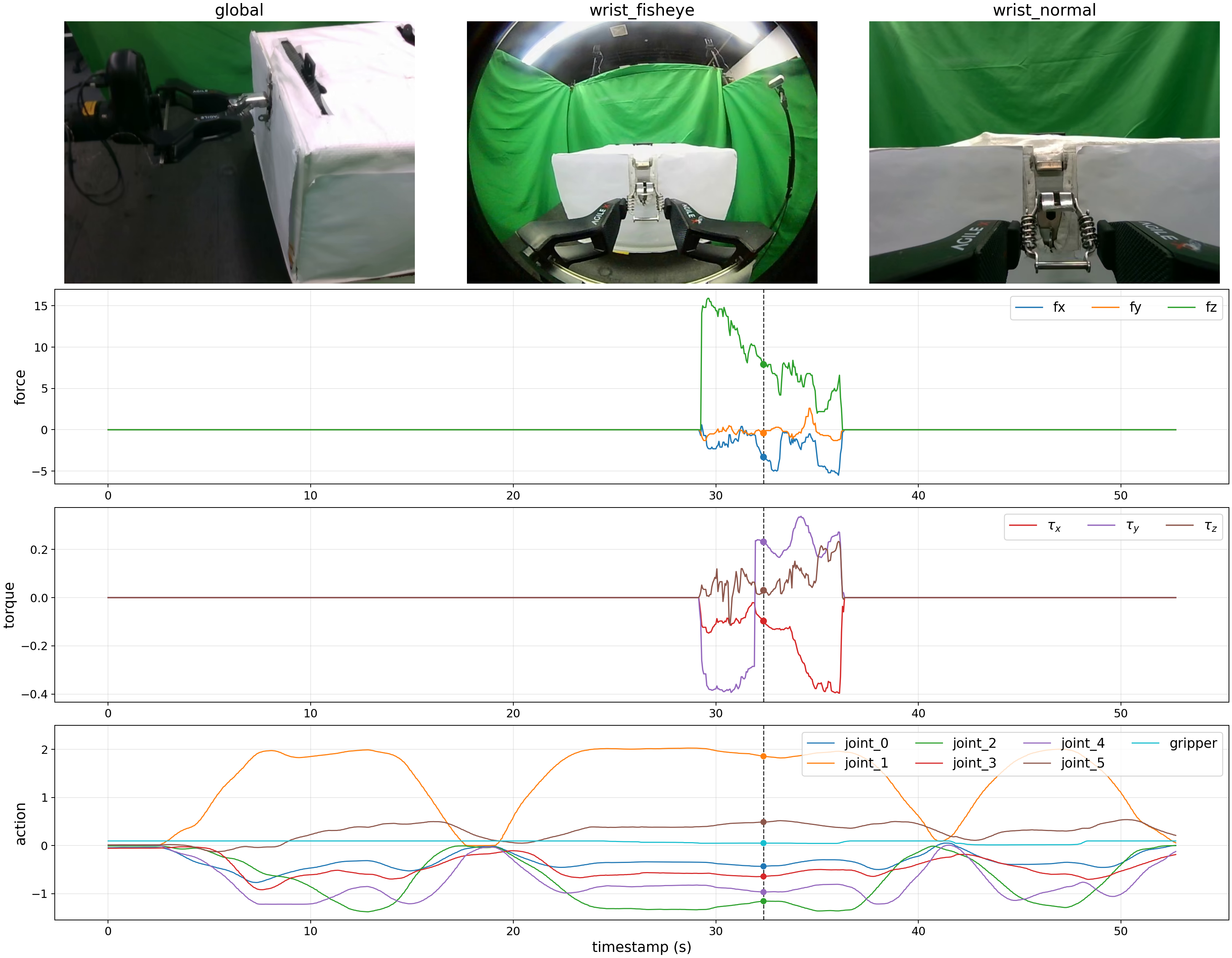} \\
\small (b) Latch-flipping phase
\end{tabular}
\caption{
Measured contact evolution in the latch-locking task. Each panel shows synchronized camera views, the measured wrench $(f_x, f_y, f_z, \tau_x, \tau_y, \tau_z)$ from the right-fingertip tactile sensor, and action traces. The sequence covers the process from initially grasping the latch to flipping it.
}
\label{fig:app_measured_wrench_latch}
\end{figure*}

\begin{figure*}[t]
\centering
\setlength{\tabcolsep}{3pt}
\setlength{\abovecaptionskip}{3pt}
\setlength{\belowcaptionskip}{-3pt}

\begin{tabular}{cc}
\begin{minipage}[t]{0.48\textwidth}
\centering
\includegraphics[width=\linewidth]{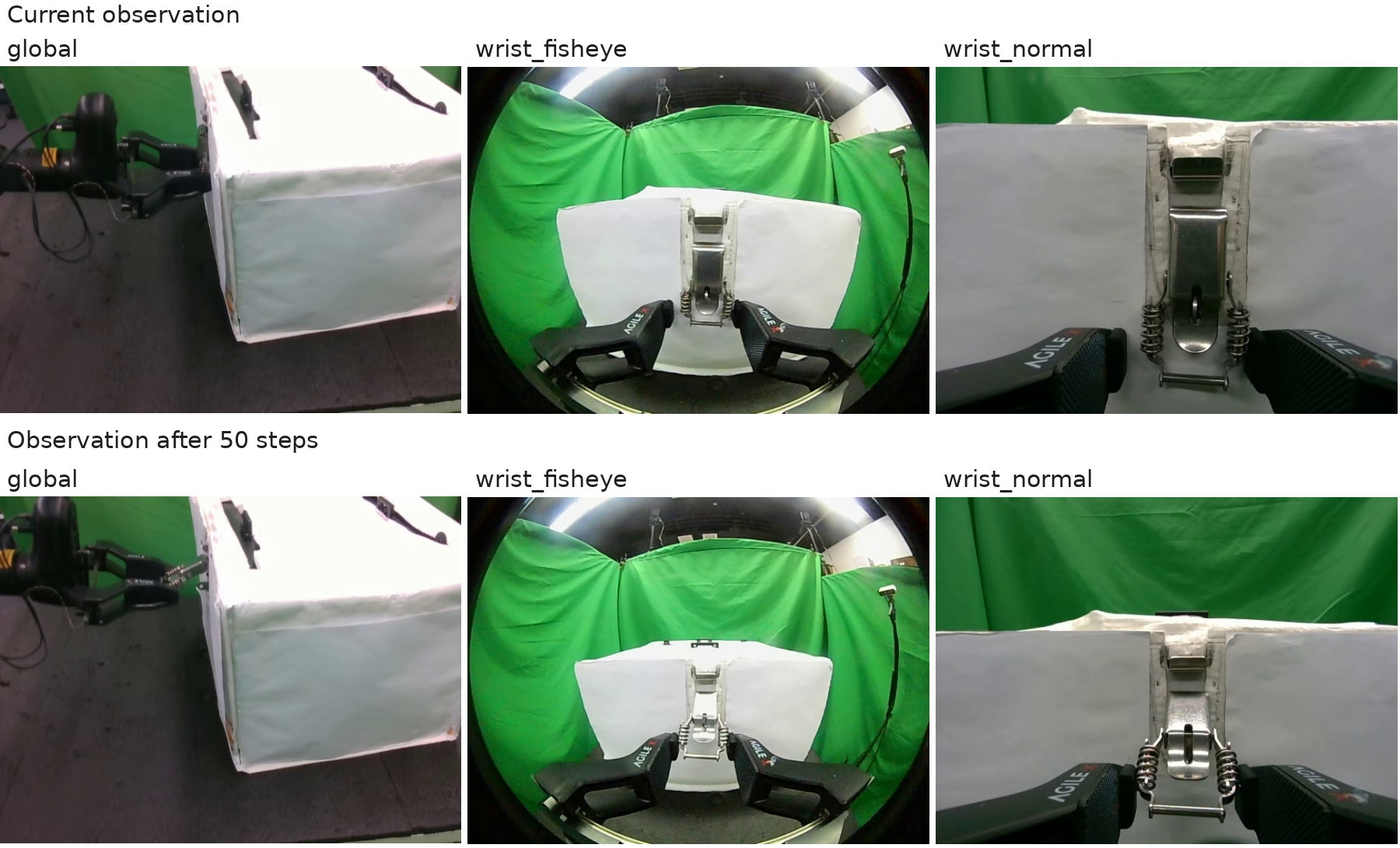}\\
\vspace{-0.1em}
\includegraphics[width=\linewidth]{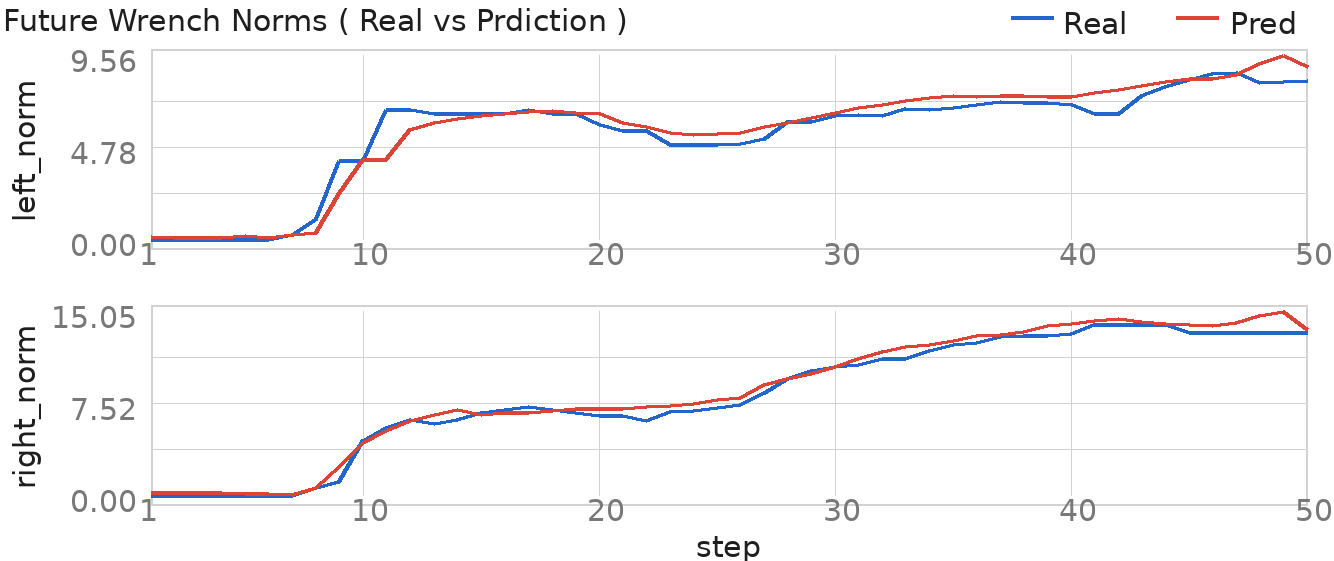}\\
\footnotesize (a1) Latch task: beginning of locking
\end{minipage}
&
\begin{minipage}[t]{0.48\textwidth}
\centering
\includegraphics[width=\linewidth]{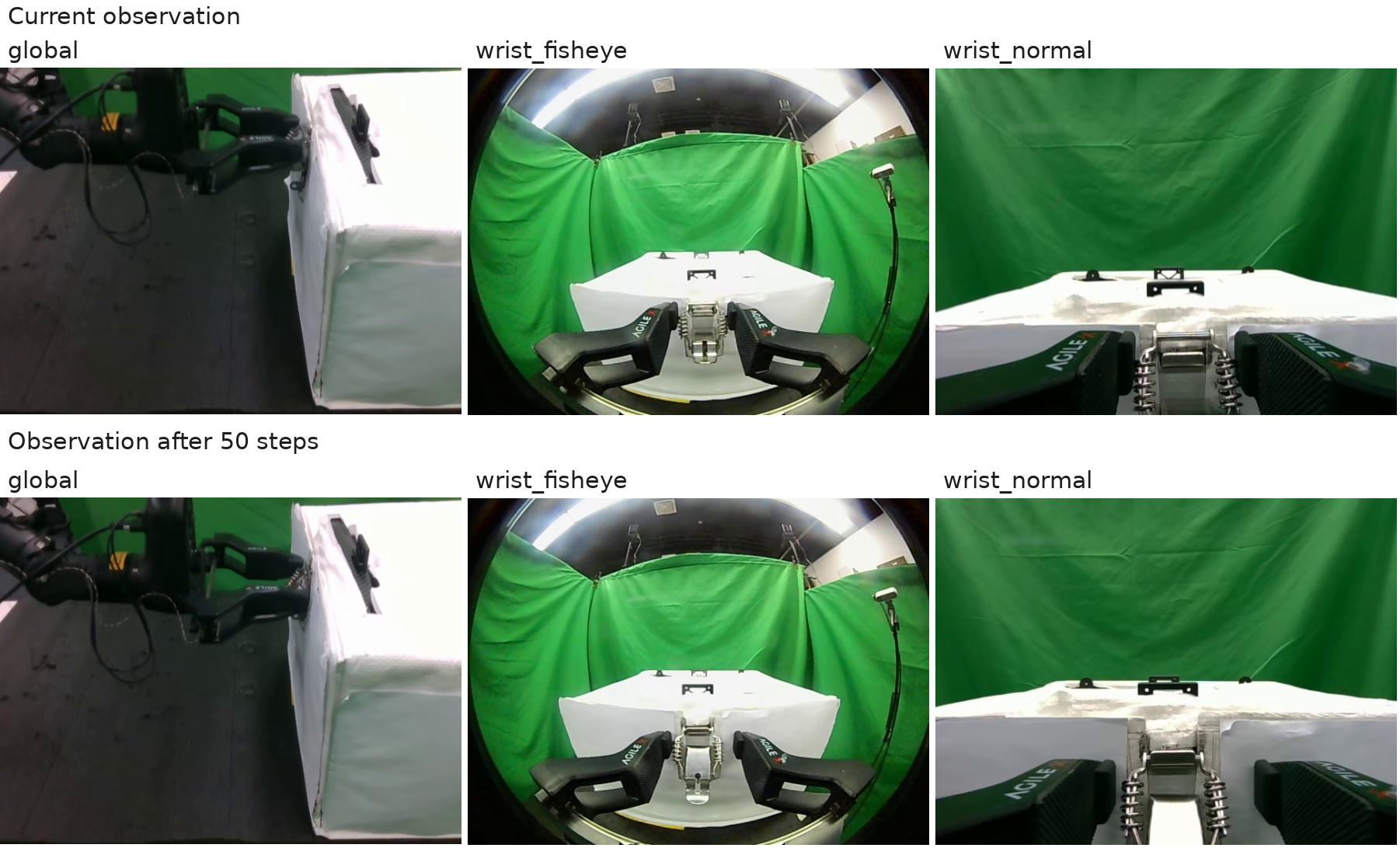}\\
\vspace{-0.1em}
\includegraphics[width=\linewidth]{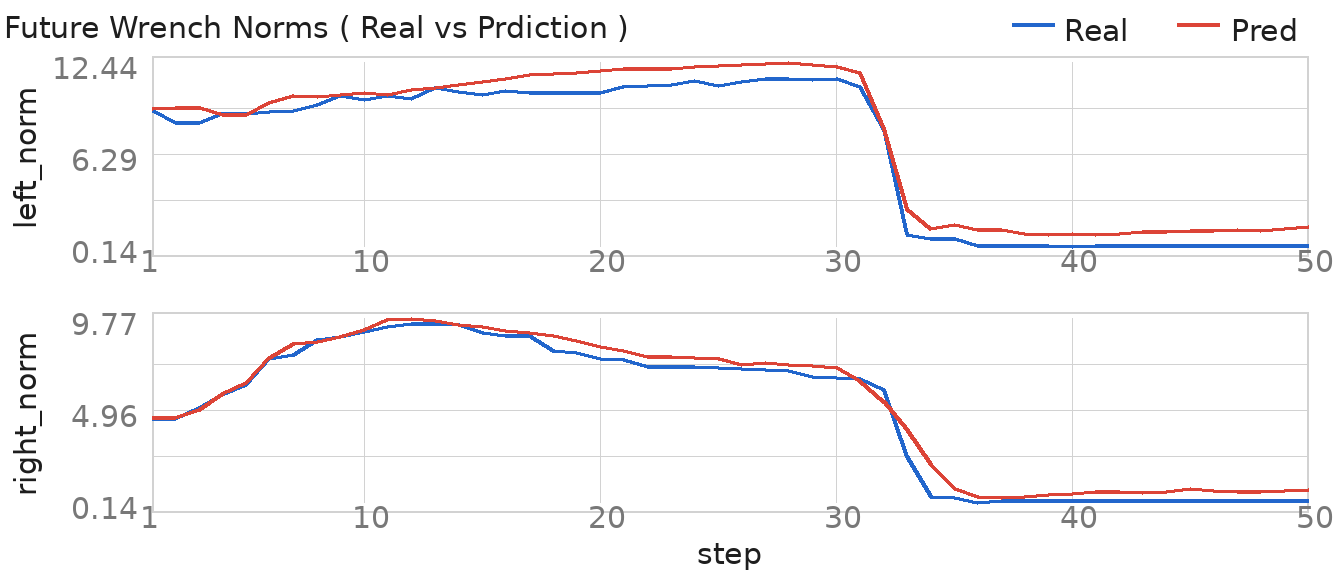}\\
\footnotesize (a2) Latch task: end of locking
\end{minipage}
\\[0.2em]

\begin{minipage}[t]{0.48\textwidth}
\centering
\includegraphics[width=\linewidth]{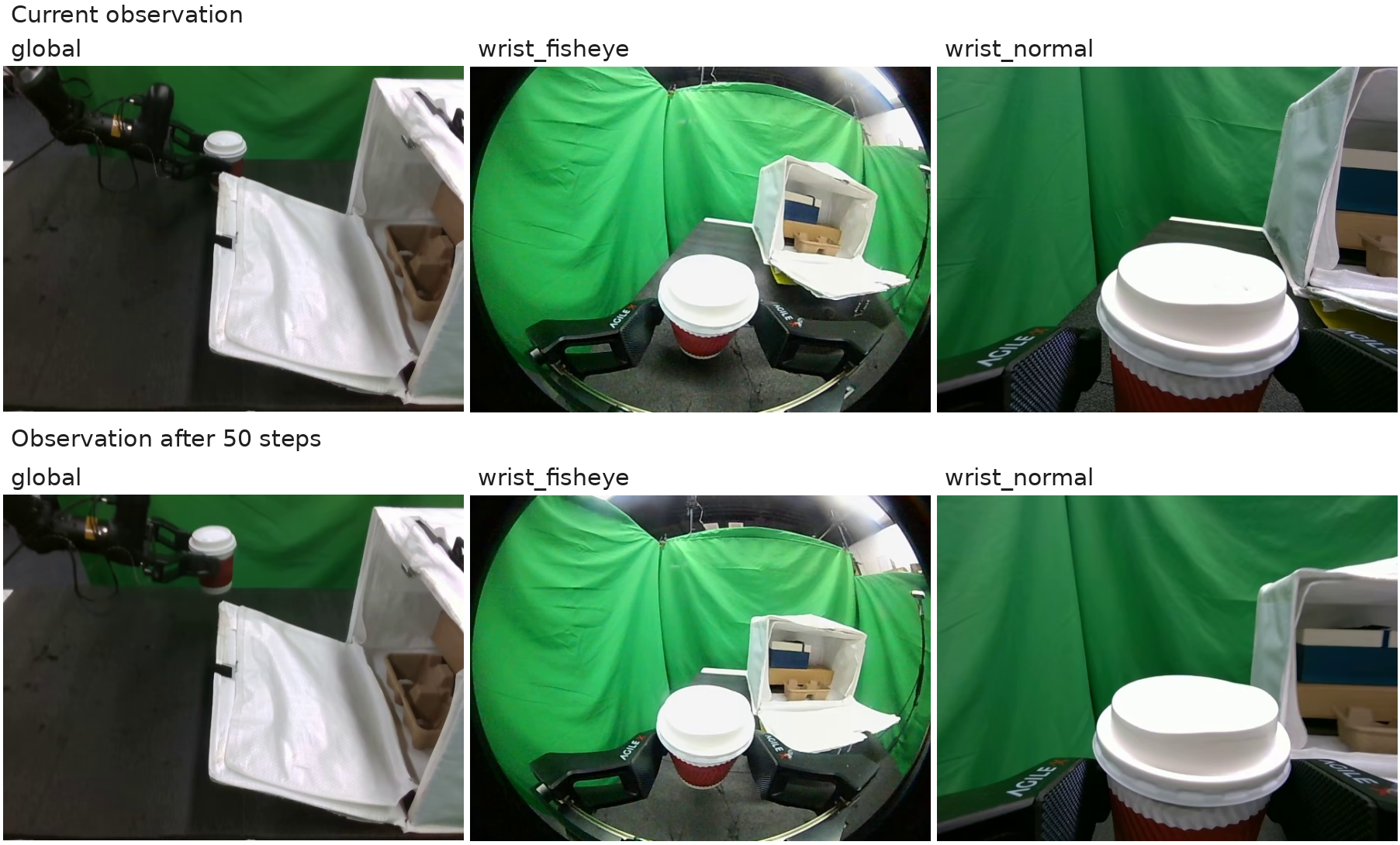}\\
\vspace{-0.1em}
\includegraphics[width=\linewidth]{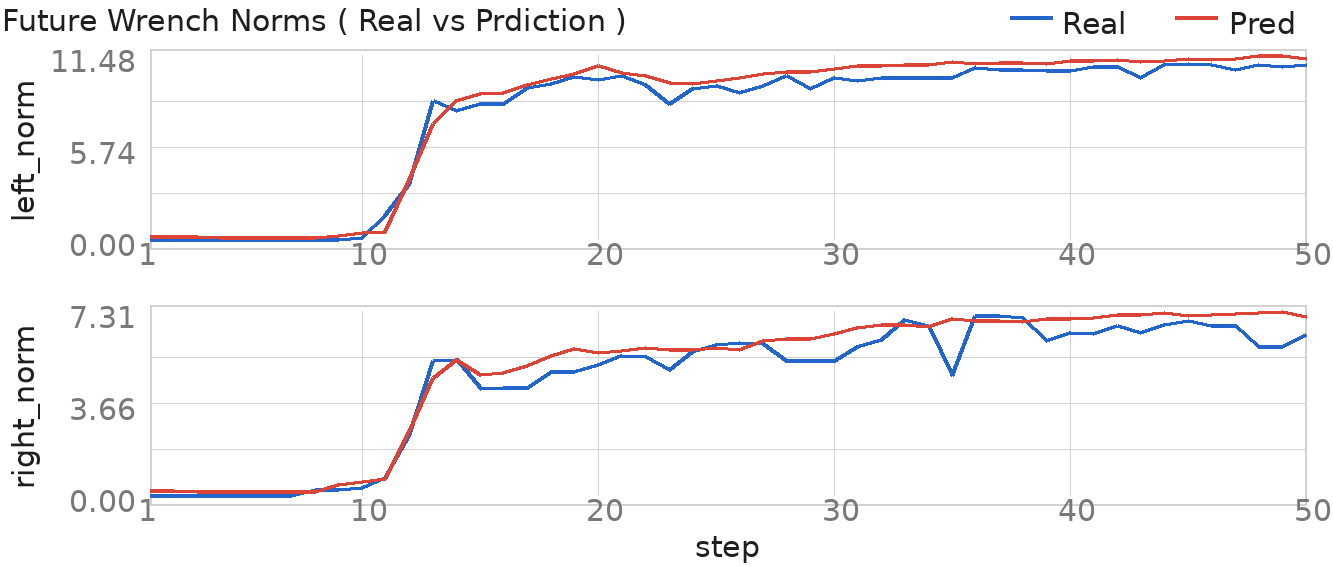}\\
\footnotesize (b1) Cup task: cup grasping
\end{minipage}
&
\begin{minipage}[t]{0.48\textwidth}
\centering
\includegraphics[width=\linewidth]{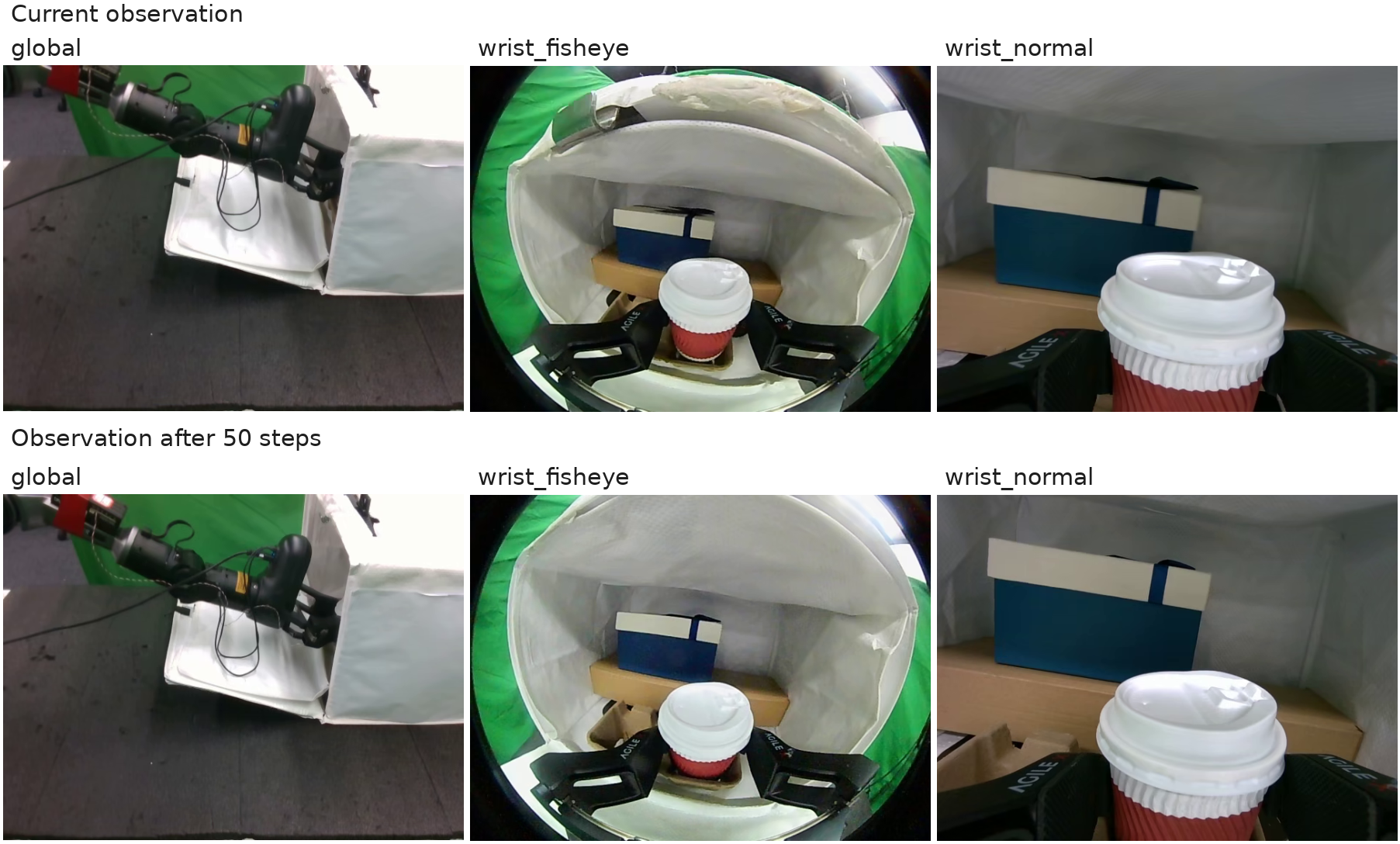}\\
\vspace{-0.1em}
\includegraphics[width=\linewidth]{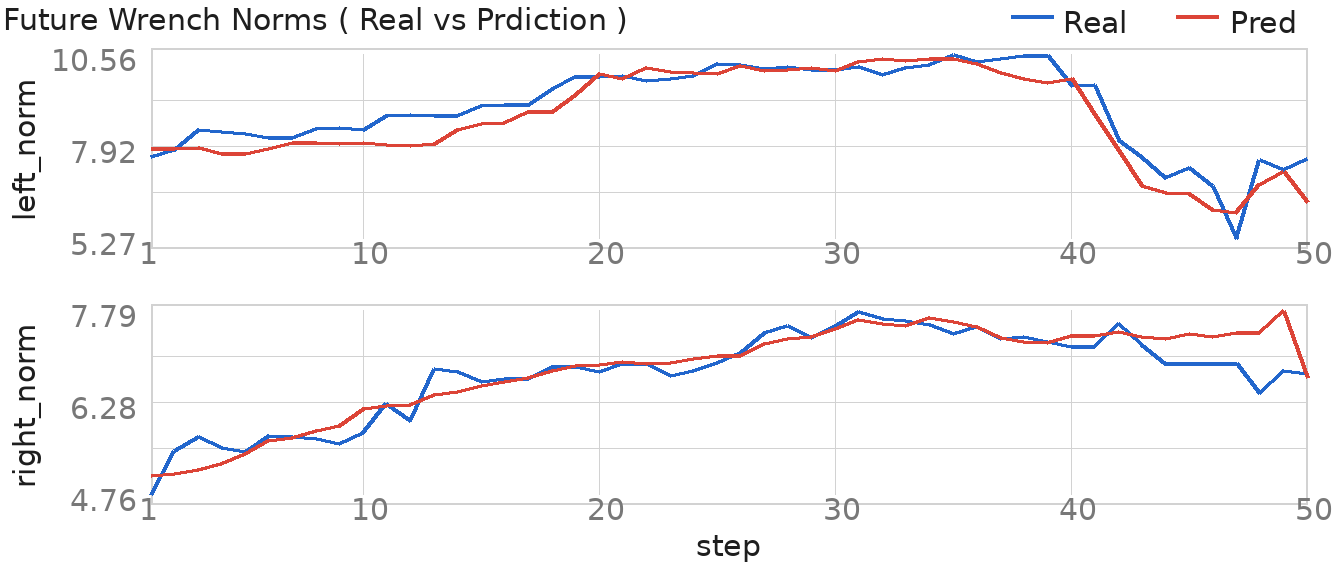}\\
\footnotesize (b2) Cup task: insertion into the holder
\end{minipage}
\\[0.2em]

\begin{minipage}[t]{0.48\textwidth}
\centering
\includegraphics[width=\linewidth]{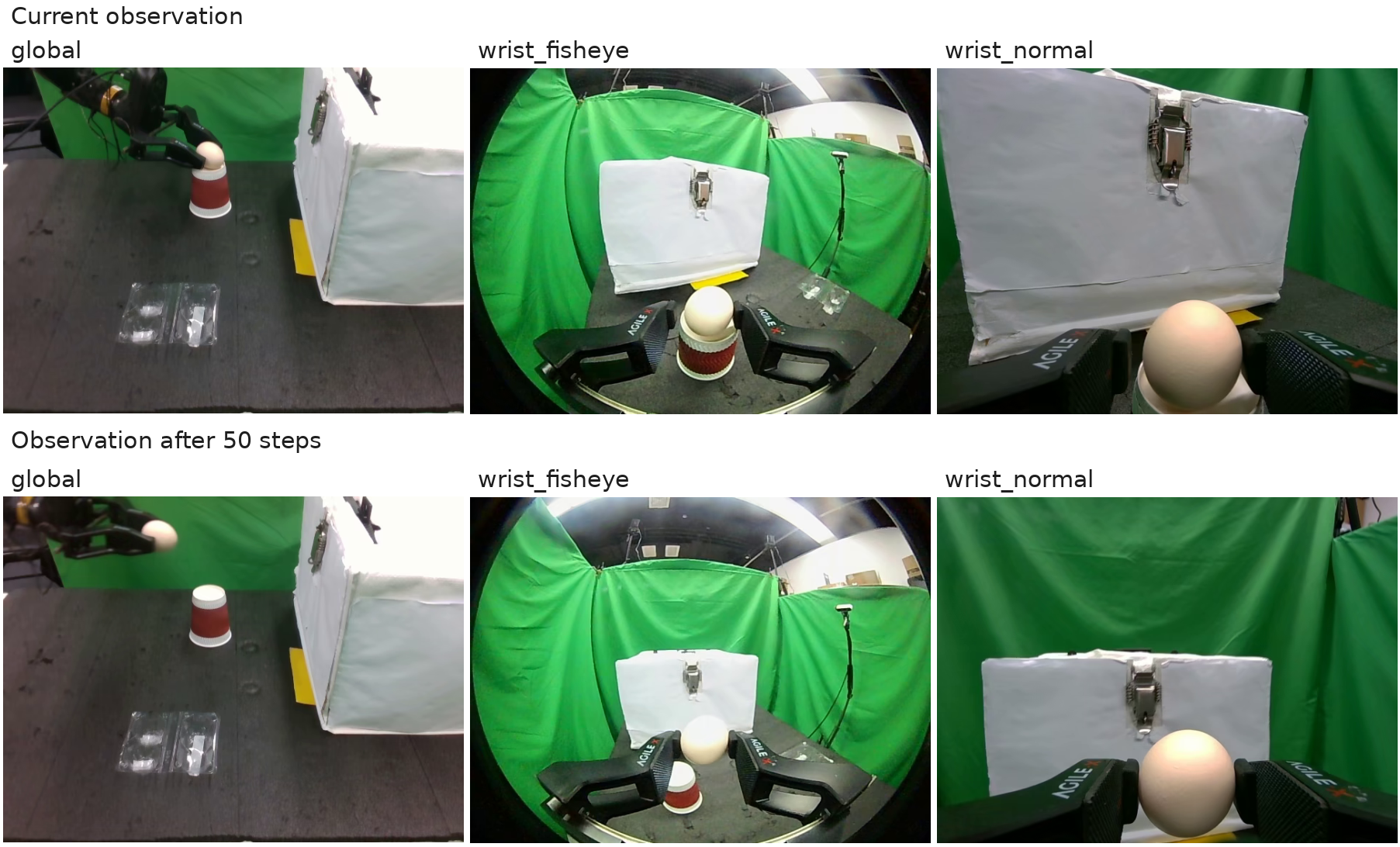}\\
\vspace{-0.1em}
\includegraphics[width=\linewidth]{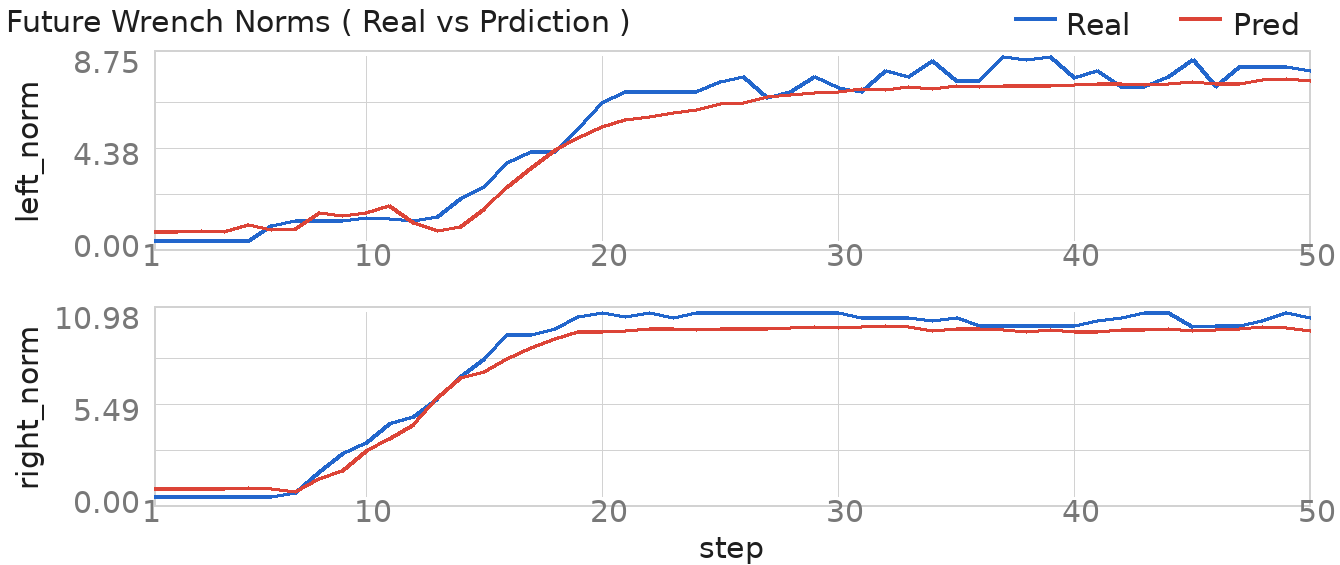}\\
\footnotesize (c1) Egg task: beginning of grasping
\end{minipage}
&
\begin{minipage}[t]{0.48\textwidth}
\centering
\includegraphics[width=\linewidth]{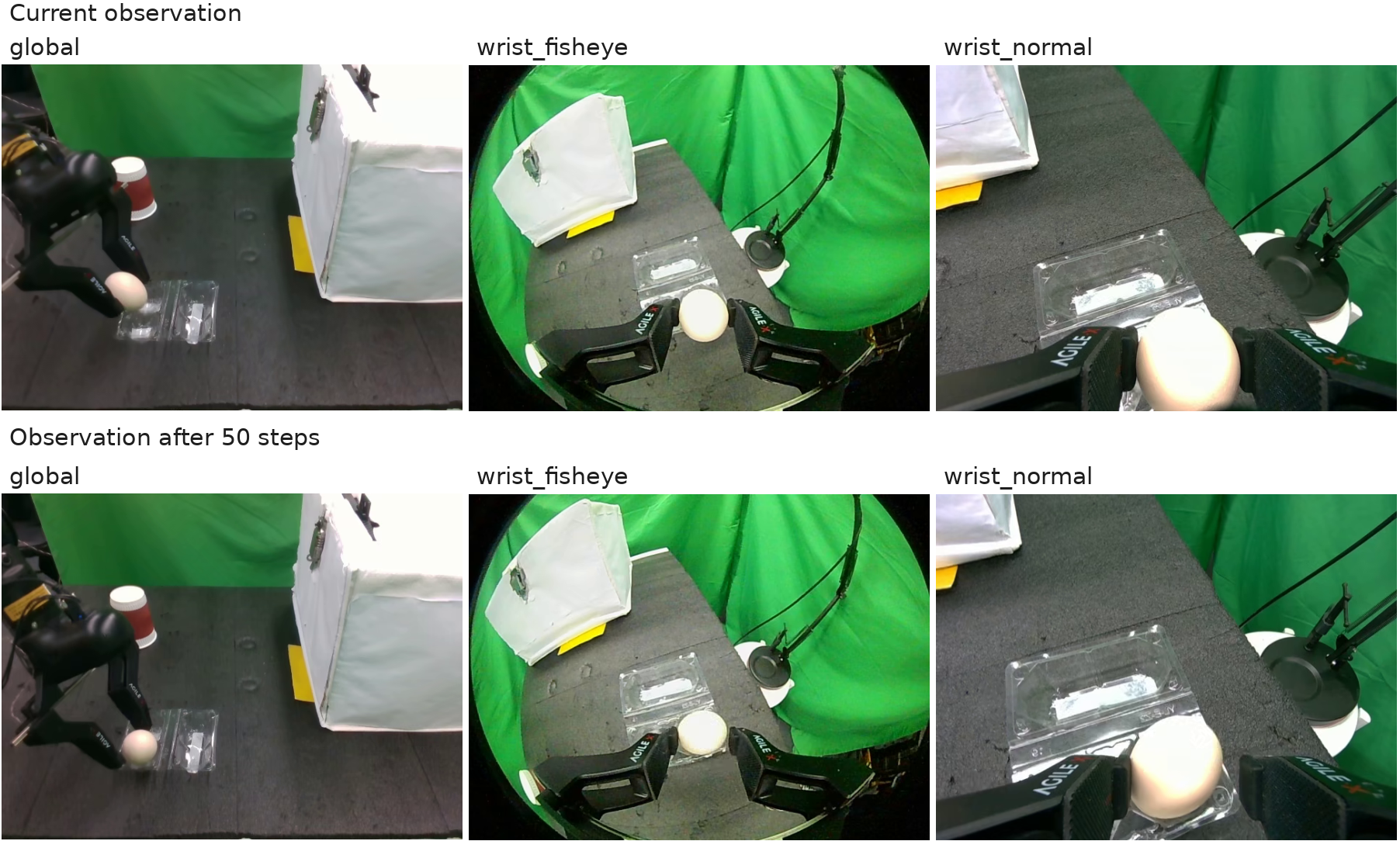}\\
\vspace{-0.1em}
\includegraphics[width=\linewidth]{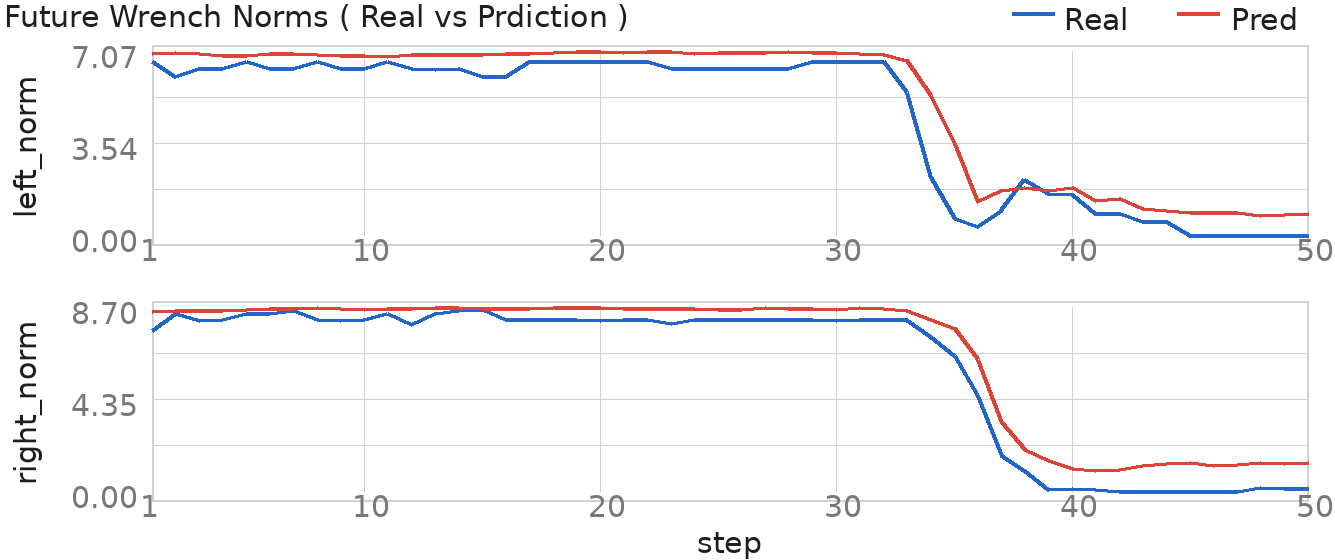}\\
\footnotesize (c2) Egg task: end of grasping
\end{minipage}
\end{tabular}

\caption{
Future-wrench prediction examples across the latch, cup, and egg tasks. Each panel shows the current and post-50-step observations above, and compares the predicted future wrench norms with the measured wrench norms from the left and right fingertip tactile sensors below.
}
\label{fig:app_future_wrench_examples_all}
\end{figure*}

\end{document}